\documentclass{article}
\usepackage[final]{staix_2026}  

\usepackage[utf8]{inputenc}
\usepackage[T1]{fontenc}
\usepackage{hyperref}
\usepackage{url}
\usepackage{booktabs}
\usepackage{amsfonts}
\usepackage{nicefrac}
\usepackage{microtype}
\usepackage{xcolor}
\usepackage{graphicx}

\RequirePackage{amsthm,amsmath,amsfonts,amssymb}
\usepackage{float}
\usepackage{babel}
\usepackage{amsmath}
\usepackage{bbm}
\usepackage{algorithm}
\usepackage{multicol}

\theoremstyle{plain}

\newtheorem{theorem}{Theorem}[section]
\newtheorem{lemma}[theorem]{Lemma}

\newtheorem{remark}[theorem]{Remark}

\theoremstyle{remark}

\def \bx{\mathbf{x}}

\def \be{\begin{align*}}
\def \ee{\end{align*}}

\def \E{\mathbb{E}}

\def \R{\mathbb{R}}
\def \X{\mathbb{X}}

\title{Cross-Cluster Weighted Forests}

\author{%
  Maya Ramchandran \\
  Department of Biostatistics, Harvard T.H. Chan School of Public Health \\
  \texttt{ramchandran.maya@gmail.com}
  \And
  Rajarshi Mukherjee \\
  Department of Biostatistics, Harvard T.H. Chan School of Public Health \\
   \And
  Giovanni Parmigiani \\
  Department of Biostatistics, Harvard T.H. Chan School of Public Health \\
  Department of Data Science, Dana-Farber Cancer Institute \\
}

\begin{document}

\maketitle

\begin{abstract}
Building trustworthy machine learning algorithms for biological applications requires adapting to data heterogeneity from different sources, batches, distributions, or studies. We propose the \textit{Cross-Cluster Weighted Forest} (CCWF), an ensembling approach that explicitly leverages heterogeneity in the feature distribution to produce more accurate and more generalizable predictors than the standard Random Forest in cases when data can be naturally clustered. CCWF generalizes the RF architecture to an outer unsupervised layer, supervised subtasks, and ensembling. Specifically it involves unsupervised clustering of the training data, fitting a Random Forest on each cluster, and combining the forests via stacked regression weights that reward cross-cluster generalizability. We provide a theoretical analysis of an analytically tractable forest model showing that cluster-based ensembling is asymptotically more accurate than training a single forest on the full data, with the gain driven by bias reduction. In simulations, we find that CCWF is robust across data-generating regimes and outcome models; furthermore, we explore the influence of data partitioning and ensemble weighting strategies on the benefits of our method. Finally, we apply our approach to cancer molecular profiling and gene expression datasets that are naturally divisible into clusters; in both simulations and real data examples, we illustrate that our approach outperforms classic Random Forest by margins of 30-40\%, aligning with our theoretical results. Overall, we show that CCWF provides a statistically grounded prediction algorithm for data spanning multiple domains or sub-populations, a structure common in biological applications.
\end{abstract}

\section{Introduction}
Natural clusters and batch effects are common in biological datasets pooled across studies, sites, instruments, or biological sub-populations. This heterogeneity poses a central challenge for trustworthy and generalizable machine learning: a model that is accurate on the full training distribution can fail unpredictably on a single sub-population or new domain \citep{goh2017batch, chauhan2010data, leek2010tackling, johnson2007adjusting, tran2020benchmark}. A growing body of work addresses this phenomenon through mixed-effects models, mixture-of-experts architectures, partition-then-ensemble strategies, and clustered federated learning, all of which reflect the principle that learning across domains is most reliable when algorithms explicitly account for heterogeneity rather than averaging it away \citep{luo2010, verbeke1996, Dietterich2000, Bouwmeester2013, jacobs1991adaptive, jordan1994hierarchical, fedus2022switch, deodhar2007, Trivedi2015, Ramchandran2020, strehl2002cluster, ghosh2022efficient, sattler2021clustered}.

Random Forest (RF) represent a highly popular tabular data algorithm and are extensively used in genomic and biomedical applications \citep{Patil2018, Sharkey1996, randomForest, Ramchandran2020, bernau2014cross, breiman2001, Biau2016}. Theoretical work has characterized their consistency, convergence rates, and high-dimensional behavior \citep{scornet2015consistency, mourtada2020minimax, chi2022asymptotic}. Empirically, \cite{Ramchandran2020} showed that ensembles of RFs trained on heterogeneous datasets can outperform a single forest trained on all available data, aligning with the multi-study ensembling approach of \cite{Patil2018}, in which learners trained on separate studies are stacked to improve cross-study performance. These findings suggest that, for datasets with cluster-like heterogeneity, partitioning data before ensembling forests may improve over the classic RF algorithm.

We evaluate this hypothesis when the outcome model is shared across clusters and the primary heterogeneity lies in the covariate distribution. This setting extends the classic covariate shift paradigm by considering internal heterogeneity within the training distribution itself \citep{sugiyama2007covariate, shimodaira2000improving, quinonero2009dataset}. It is also closely related to domain generalization and distributionally robust learning, which seek predictors that maintain performance across heterogeneous training and test populations \citep{koh2021wilds, sagawa2020distributionally, wang2022generalizing}. Throughout the paper, we distinguish between `estimated clusters' obtained algorithmically and `true clusters' arising from the data-generating process.

We introduce the Cross-Cluster Weighted Forest (CCWF), a learning approach designed to improve prediction performance by leveraging covariate heterogeneity. CCWF first clusters the training data using algorithms such as k-means, then trains separate RFs within each estimated cluster, and finally uses stacked regression to assign weights to these forests based on their stability and predictive accuracy across clusters \citep{Stacking1996, wolpert1992stacked, vanderlaan2007super}. This strategy reduces within-cluster heterogeneity while preserving cross-cluster learner diversity, following the ensemble learning principle that combining diverse learners can improve generalization \citep{Schapire2003, Dietterich2000, Bagging1996, kuncheva2003measures}. We study both the theoretical and empirical properties of CCWF, demonstrating its ability to outperform standard RF in simulated settings and real biological datasets with natural cluster structure.

Our approach builds on prior work using clustering to partition data before training ensemble learners, including methods based on simultaneous co-clustering and model fitting or clustering as a preprocessing step \citep{deodhar2007, Trivedi2015, strehl2002cluster}. CCWF differs by using generic clustering algorithms to identify substructure and cross-cluster weighting to favor learners that generalize across partitions. It also differs from prior work linking RFs to clustering, which typically uses RF itself as a clustering method rather than using external cluster structure to form weighted forest ensembles \citep{Yan2013, Shi2006, Bicego2019}. Finally, it differs from \citep{young2025clustered}, who develop an ensemble of RF's method for data with independent groups that each have a known dependence structure, such as repeated measures/longitudinal data. The method's primary goal is to achieve better precision by accounting for this known correlation structure within groups.

The remainder of the paper is organized as follows. We first present a theoretical analysis of CCWF using an analytically tractable formulation to study how bias--variance tradeoffs determine when cluster-based ensembling improves prediction on heterogeneous data. We then validate CCWF empirically through experiments on simulated datasets with cluster structure, examining the impact of each algorithmic step and assessing robustness across data-generating scenarios. Finally, we apply CCWF to real gene expression and clinical datasets with feature-distribution heterogeneity, showing improved performance relative to classic RF.

CCWF is a structural generalization of the RF architecture using an outer unsupervised layer, supervised subtasks, and ensembling. While in theory it can be compared to any tree-based algorithm such as  gradient-boosted trees or deep tabular models, we benchmark against baselines where the supervised subtasks are the same in our algorithm and the comparators. In this way, we are able to provide direct evidence on the value of the architecture. Our baselines therefore hold the learner class fixed and vary only the partitioning and ensembling procedure, isolating whether covariate-aware clustering and weighting improve over standard RF training.

\section{Algorithm Overview}
\begin{algorithm}[t]
Set the number of clusters $k \geq 2$.
\begin{enumerate}
    \item Partition the training set into $k$ disjoint clusters using an unsupervised clustering algorithm on all or a subset of the features.
    \item For $b = 1,\ldots,k$:
    \begin{itemize}
        \item Train a Random Forest on the $b^{th}$ cluster.
        \item Compute predictions $\hat{\mathbbm{Y}}_b$ for all training points.
    \end{itemize}
    \item Compute non-negative stacked regression weights
    \[
        \hat{\boldsymbol{w}} = \arg\min_{\boldsymbol{w} \geq 0} \left\| \mathbbm{Y} - \sum_{b=1}^k w_b \hat{\mathbbm{Y}}_b \right\|_2^2 + \lambda \|\boldsymbol{w}\|_2^2,
    \]
    with $\lambda$ selected via cross-validation using {\tt glmnet} \citep{glmnet}.
    \item Return $\hat Y_{\text{CCWF}}(\bx_\star) = \sum_{b=1}^k \hat w_b\, \hat Y_b(\bx_\star)$.
\end{enumerate}
\caption{Cross-Cluster Weighted Forest (CCWF)}
\label{algo:algo1}
\end{algorithm}

The CCWF approach is outlined in Algorithm~\ref{algo:algo1}. Standard RF averages equally weighted trees trained on bootstrap samples. Our approach introduces a non-random data partitioning step in which we break the data into $k$ subsections, potentially improving performance when the data can be naturally clustered. We use k-means for scalability, though the framework permits other clustering methods \citep{bottou1995convergence, hartigan1979algorithm, coates2012learning}. We train RFs on each cluster and combine in-sample training predictions using nonnegative ridge stacking; weights are learned only from training data, and generalization is evaluated on held-out test data \citep{Patil2018, Guan2019a, Ramchandran2020, Stacking1996}.

\section{Theory}

In this section, we examine the asymptotic risk of CCWF compared to the classic RF algorithm using an analytical setup that approximates the full algorithm as closely as is possible given current theoretical results on Random Forest \citep{scornet2026theory}. We show that bias reduction is the primary mechanism through which the general cluster-based ensembling framework produces improvements for forest-based learners. 

\subsection{Notation}

We denote the training data set as $\mathcal{D}_n=(\X,\mathbbm{Y})\in\R^{n\times S}\times\R^n$, where $S$ is the number of covariates and $n$ is the training sample size. The training dataset can be further expressed as $(\X,\mathbbm{Y})=\left(\left[\X_1 \ldots \X_k\right]^T,\left[\mathbbm{Y}_1 \ldots \mathbbm{Y}_k\right]\right)$, where $k$ is the number of clusters, $\X_b\in\R^{n_b\times S}$, and $\mathbbm{Y}_b\in\R^{n_b}$ denote the covariates and outcomes for cluster $b$. 
We specify that $n_b=n/k$ and that the training covariate set consists of $k$ equally sized, non-overlapping, uniform clusters spanning the $[0,1]^S$ hypercube: $\X_1\stackrel{\rm i.i.d.}{\sim}\left[\mathrm{U}\left(0,\frac{1}{k}\right)\right]^S$, $\X_2\stackrel{\rm i.i.d.}{\sim}\left[\mathrm{U}\left(\frac{1}{k},\frac{2}{k}\right)\right]^S$, $\hdots$, $\X_k\stackrel{\rm i.i.d.}{\sim}\left[\mathrm{U}\left(\frac{k-1}{k},1\right)\right]^S$. Equivalently, each training sample is $(\bx_i,Y_i)$, where $A_i\sim\mathrm{Categorical}\left(\frac{1}{k},\ldots,\frac{1}{k}\right)$ denotes the cluster assignment, $\bx_i\mid A_i=b$ is uniform on $\left[\frac{b-1}{k},\frac{b}{k}\right]^S$, and $Y_i=f(\bx_i)$. We make no parametric assumptions on $f(\cdot)$. Test point $(\bx_\star,Y_\star)$ follows the same distribution as the training set.

\subsubsection{Modeling Approaches}
We analyze the Centered Random Forest (CRF) of \cite{Breiman2004}, which shares the recursive axis-aligned-split structure of standard RFs but uses coordinate-uniform splits independent of the response. CRFs are the standard analytical surrogate in the RF theory literature \citep{Biau2016, klusowski2020} because conclusions about bias and variance scaling carry over qualitatively to standard RFs while remaining tractable. We later evaluate whether the bias-dominance prediction holds for standard RFs (Section 4). Full definitions are in Appendix \ref{sec:centered_forest_defs}.

The \textit{Ensemble} method representing an analytically tractable formulation of CCWF functions by training a CRF on $(\X_b, \mathbbm{Y}_b)$ for each cluster $b$, yielding predictions $\hat Y_b(\bx_\star; \theta, \mathcal{D}_n)$ which are then summed to form the final \textit{Ensemble} prediction: $\hat Y_E(\bx_\star) = \sum_{b=1}^k \hat Y_b(\bx_\star)$. Because CRFs do not extrapolate beyond their training data, $\hat Y_b(\bx_\star) = 0$ if the cluster assignment of $\bx_\star \neq b$ and thus $\hat Y_E(\bx_\star)$ is in fact the prediction of the single forest whose training data range contains $\bx_\star$. This construction isolates the mechanism that CCWF is designed to exploit: when clusters are well separated and learners do not extrapolate outside their training ranges, cluster-specific forests specialize to the relevant region of covariate space. As a baseline, we train a single CRF on the full dataset termed the \textit{Merged}, with prediction $\hat Y_M(\bx_\star)$. The aim of this section is to derive the limit of $\frac{\textit{RMSE}(\hat Y_E)}{\textit{RMSE}(\hat Y_M)} = \frac{\sqrt{\mathbbm{B}_E^2 + \mathbbm{V}_E}}{\sqrt{\mathbbm{B}_M^2 + \mathbbm{V}_M}}$, where $\mathbbm{B}, \mathbbm{V}$ are the bias and variance of the \textit{Ensemble} and \textit{Merged}.

\subsection{Asymptotic Analysis}

In Lemma \ref{lemma:forestpredictions} presented in the Appendix, we show that the \textit{Ensemble} and \textit{Merged} predictions can be represented as weighted averages of the training outcome values, with the choice of weights as the differentiating factor between the methods. To develop theoretical results regarding the performance of these two approaches, we build on \cite{klusowski2020} in characterizing the asymptotic behavior of the bias and variance components of the MSE for CRFs. We extend upon these results to allow for clusters in the training and test data and to remove any requirement on the functional form of the outcome model $f(\bx)$. We note that the latter represents a significant step forward upon previous analytical work on Random Forests, which typically assume a linear or piecewise linear outcome model. 

We begin by stating supplementary results that support the primary performance comparison between the \textit{Ensemble} and the \textit{Merged} in Theorem \ref{theorem: rf_uniform} below. First, we can show that the variance terms $\mathbbm{V}_E$ and $\mathbbm{V}_M \to 0$ at a rate of $O\left(\frac{1}{n}\right)$, the formal statement and proof of which we provide in Lemma \ref{lemma:variancetozero} in the Appendix. This follows directly from the results in [\cite{klusowski2020}, Proposition 2] and its corresponding proof. We further show in Remark~\ref{rem: biasconvergence} in the Appendix that the bias terms $\mathbbm{B}_E$ and $\mathbbm{B}_M$ converge more slowly than $O(1/n)$ under mild assumptions on $S$ and $c_n$.
Therefore, we can characterize the limit of $\frac{\textit{RMSE}(\hat{Y}_E(\bx_{\star}))}{\textit{RMSE}(\hat{Y}_M(\bx_{\star}))}$ by the limit of $\frac{\mathbbm{B}_E}{\mathbbm{B_M}}$; that is, the bias terms dominate the RMSE as $n \to \infty$. We can then derive the limit of the performance ratio between the \textit{Ensemble} and the \textit{Merged} as presented in Theorem \ref{theorem: rf_uniform} below.   

\begin{theorem} 
\label{theorem: rf_uniform}
Given a training set with $n$ samples, $S$ covariates, and $k \geq 2$ equally-sized uniform clusters 
$\{(\bx_i,Y_i): i=1,\ldots,n\}$, where 
$A_i\sim\mathrm{Categorical}\left(\frac{1}{k},\ldots,\frac{1}{k}\right)$ denotes the cluster assignment,
$\bx_i \mid A_i=b$ is drawn uniformly from the hypercube 
$\left[\frac{b-1}{k},\frac{b}{k}\right]^S$, and $Y_i=f(\bx_i)$, 
we can represent the predictions of the \textit{Ensemble} and the \textit{Merged} on test point 
$(\bx_\star,Y_\star)$ drawn from the same distribution, as $\hat{Y}_E$ and $\hat{Y}_M$ respectively. Then,
\begin{align*}
\lim_{n \to \infty} \frac{\textit{RMSE}(\hat{Y}_E(\bx_{\star}))}{\textit{RMSE}(\hat{Y}_M(\bx_{\star}))} 
&= \frac{1}{\sqrt{2}} .
\end{align*} 
\end{theorem}

Theorem \ref{theorem: rf_uniform} demonstrates that the asymptotic performance improvement of the \textit{Ensemble} over the \textit{Merged} is independent of the number of true clusters $k$ for $k \geq 2$. To understand this phenomenon, consider that ensembling involves a trade-off: it reduces the number of samples used to train each learner (decreasing efficiency compared to the \textit{Merged}), but it also allows each learner to focus on regions of the covariate space that better reflect the distribution of the test point, thereby improving accuracy. The proof of Theorem \ref{theorem: rf_uniform} decomposes the ratio $\frac{\textit{RMSE}(\hat{Y}_E(\bx{\star}))}{\textit{RMSE}(\hat{Y}_M(\bx{\star}))}$ into two components that capture this trade-off. It shows that when these components are multiplied, terms involving $k$ cancel out, resulting in a final ratio of $\frac{1}{\sqrt{2}}$. Although each learner in the \textit{Ensemble} is trained on $\frac{1}{k}$ times the number of samples as the \textit{Merged}, the benefit of the \textit{Ensemble} approach lies in its ability to more effectively parse out the heterogeneity introduced by clusters from the true covariate-outcome relationship. This advantage outweighs the loss in efficiency due to the reduced sample size as $n \to \infty$. This result provides a theoretical characterization of how covariate heterogeneity can affect forest ensembling without imposing a parametric form on $f(\bx)$. We verify that the ratio in Theorem \ref{theorem: rf_uniform} holds exactly in a simulation study presented in Appendix section \ref{sec:sim_study_theory}. 

\section{Simulations: Ablation and Sensitivity Experiments}
\subsection{Terminology and Setup}

\par This section evaluates ensembling approaches on simulated data designed to mimic real genomic datasets. We first examine the  importance of data partitioning on overall results, highlighting the benefits of clustering algorithms. Next, we analyze the impact of ensemble weighting strategies, comparing stacked regression weights to simple averaging. All results are averaged over 250 simulations per scenario (run on 4 CPU cores), with 95\% confidence bands computed as mean $\pm 1.96$ standard errors. We interchangeably refer to CCWF as the \textit{Cluster} method for clarity, as this naming aligns more naturally with the other methods we compare it to. We then evaluate it alongside three variations of Algorithm \ref{algo:algo1} that ablate various steps, defined as follows. The \textit{Random} method uses equally sized random partitions instead of clusters, keeping the number of total learners equal to $k$. The \textit{Multi} method trains RFs on true clusters (known in simulation but unavailable in practice) and combines them using stacking weights, thus named for its equivalence to multi-study stacking \citep{Patil2018}. The \textit{Merged} method serves as the baseline, training a single RF on the entire dataset. To ensure fair comparison, all methods are trained using the same number of total trees in order to isolate each algorithmic step's contribution to the performance. We use 100 trees per component forest for ensemble methods and $100k$ trees for the \textit{Merged}.

\par We employ two methods to generate clustered datasets for our simulations: multivariate Gaussian mixture models using the R package {\tt clusterGeneration} \citep{clusterGeneration}, and plasmodes more closely mimicking authentic biological clusters using the `monte' function in the R package {\tt fungible} \citep{fungible}. For each simulation per setting, we generate training data with \textit{ntrain} clusters (baseline: 5) and \textit{ntest} test datasets with 2 clusters each (baseline: 5), all containing \textit{ncoef} covariates (baseline: 20). We create response models by randomly selecting 10 covariates, drawing coefficients uniformly from $[-5,-0.5] \cup [0.5, 5]$, adding Gaussian noise, and considering non-linear relationships. Each cluster contains 500 samples at baseline, with between-cluster coefficient perturbation drawn uniformly from [0, 0.25]. Cluster separation is set at median values, allowing some overlap while remaining distinguishable. This setup enables comprehensive evaluation of feature distribution heterogeneity effects on ensemble performance across various scenarios. All modifications to this baseline strategy are outlined in the relevant results sections. 

We train Random Forests using the R package {\tt randomForest} \citep{liaw2002classification} with the number of trees as described above. We keep 'mtry' and 'nodesize' hyperparameters, representing the number of predictors randomly sampled as candidates at each split and the minimum size of terminal nodes respectively, at their default values of $\sqrt{S}$ and $1$. For stacked regression weights, we tune the ridge regression regularization parameter through 5-fold cross-validation within the training set. In experiments where we choose one optimal value of $k$ for k-means rather than display results across a variety of $k$ values, we choose this optimal value through maximizing the silhouette score over candidate $k$'s ranging from 2 to $(ntrain * 500) // 30$. 

\paragraph{Computational Efficiency.} 
For fixed total tree count $T$, CCWF with k-means can be more efficient than the \textit{Merged} baseline. Stacking and k-means add minimal overhead in comparison to forest training.
CCWF trains $k$ forests on $n/k$ samples, reducing total training complexity to $O(\frac{TSn}{k} \log \frac{n}{k}) < O(TSn \log n)$.

\subsection{The Importance of Data Partitioning}

\par
\begin{figure*}[t!]
  \centering
  \includegraphics[width=\linewidth]{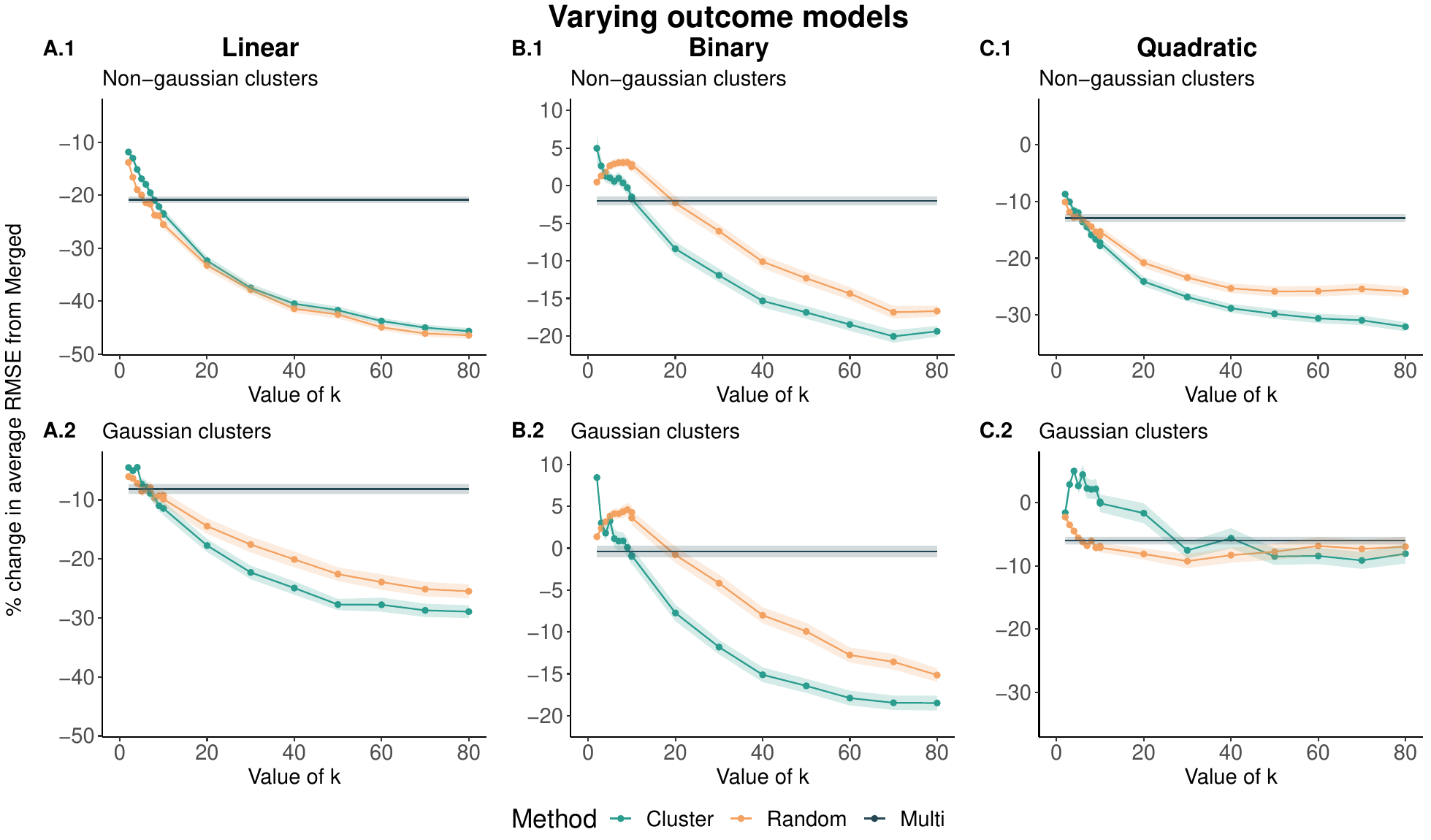}
  \caption{\it Percent change in average RMSE of ensembling approaches (color labeled) compared to the Merged across different data-generating scenarios, as a function of $k$. The first row depicts results using the non-gaussian cluster simulation approach (using the `monte' function in the {\tt fungible} package), while the second row uses a gaussian data generating model. \textbf{(A.1-A.2)} A linear model was used to generate the outcome from the covariates. \textbf{(B.1-B.2)} The binary outcome was created by using a cutoff from the linear model to create a binary step function. \textbf{(C.1-C.2)} Quadratic terms for two of the variables were added to the linear outcome-generating model. }
  \label{fig:F2}
\end{figure*}

 We investigate how data partitioning methods affect ensemble accuracy by comparing the approaches outlined in the previous section. Both the \textit{Cluster} and \textit{Random} methods train component forests on the same number of partitions, but differ in how those partitions are determined. To keep notation consistent, we refer to $k$ as the parameter used to train k-means, reflecting the number of partitions of the data also used for the \textit{Cluster} and \textit{Random} approaches. We explore the impact of varying $k$ and the data-generating mechanism across six scenarios in  Figure \ref{fig:F2}.

\par We see that the \textit{Cluster} and \textit{Random} outperform the \textit{Multi} and \textit{Merged} when $k$ exceeds the true number of clusters, with improvements over 20\% at optimal values. The \textit{Multi}, which uses true clusters, shows inconsistent performance relative to \textit{Merged}. The \textit{Cluster} and \textit{Random} are robust across outcome models and data distributions; however, the \textit{Cluster} generally outperforms the \textit{Random}, suggesting k-means clustering is preferable to random partitioning. Both methods improve as $k$ increases, plateauing when each forest trains on 35--40 out of 2500 observations. Correspondingly, average tree depth decreases, indicating that high $k$ values effectively ensemble locally weak learners akin to local smoothing approaches. This supposition is explored further in Appendix section \ref{sec:single_covariate}. Note that the theoretical result equates to a $29\%$ improvement of ensembling over merging in a more idealized setting, which is squarely within the range of improvements we see empirically. For further comparison of how the theoretical results translate in practice, we compare the \textit{Merged} and \textit{Multi} approaches and confirm that bias reduction remains the primary mechanism for performance improvements of the \textit{Multi}; a full analysis can be found in Appendix section \ref{sec:biasvar_multi_merged}.

\par The differences between the \textit{Cluster} and the \textit{Random} reveal key advantages of clustering as a partitioning strategy. Increasing the number of partitions generally improves ensemble performance, with the \textit{Multi} often outperforming other methods when $k$ is smaller than the true number of clusters. Clustering allows learners to be less correlated and more dissimilar, a property effective in other ensemble approaches. The improvement of the \textit{Cluster} over the \textit{Merged} suggests that, in these heterogeneous settings, clustering creates useful partitions for training component trees. While clustering may reduce accuracy in high-density regions, this loss does not outweigh the benefits.

\paragraph{Estimated clusters outperform true clusters.} A central and initially counterintuitive finding: even when $k$ matches the true cluster count, k-means partitions do not recover the true clusters, and yet they yield better ensembles than partitioning by ground truth. We hypothesize the mechanism: k-means minimizes within-cluster covariate range, tightening the local function approximation each forest must learn. True clusters, defined by data-generating provenance, may not minimize covariate range and often span the same covariate region as other clusters. Empirically, k-means partitioning achieves a lower average covariate range (\textbf{3.17} CI:(3.08, 3.26)) than random partitioning (\textbf{4.22} CI:(4.13, 4.31)), while true-cluster partitioning gives a much larger range (\textbf{7.33} CI:(7.21, 7.45)). By restricting each forest to a smaller subset of the covariate space, learners can more precisely discern localized patterns, especially in distribution tails. The same phenomenon persists in multi-study settings (Appendix \ref{sec:multistudy}). These results suggest that for downstream RF ensembles, \textit{covariate-range minimization} may be more useful than provenance recovery.

\paragraph{Practical $k$.} 
Silhouette analysis offers a practical diagnostic: \textit{Cluster} outperforms \textit{Merged} only when silhouette analysis selects an optimal $k > 1$, regardless of whether true clusters exist or how separated they are. Similar behavior holds for other clustering methods, so the unsupervised clustering step itself is a reasonable indicator of downstream gains. 

\subsection{Sensitivity Analyses to Realistic Dataset Variety}
\label{sec:realistic_dataset_variety}

\begin{figure*}[h]
  \centering
  \includegraphics[width=.8\linewidth]{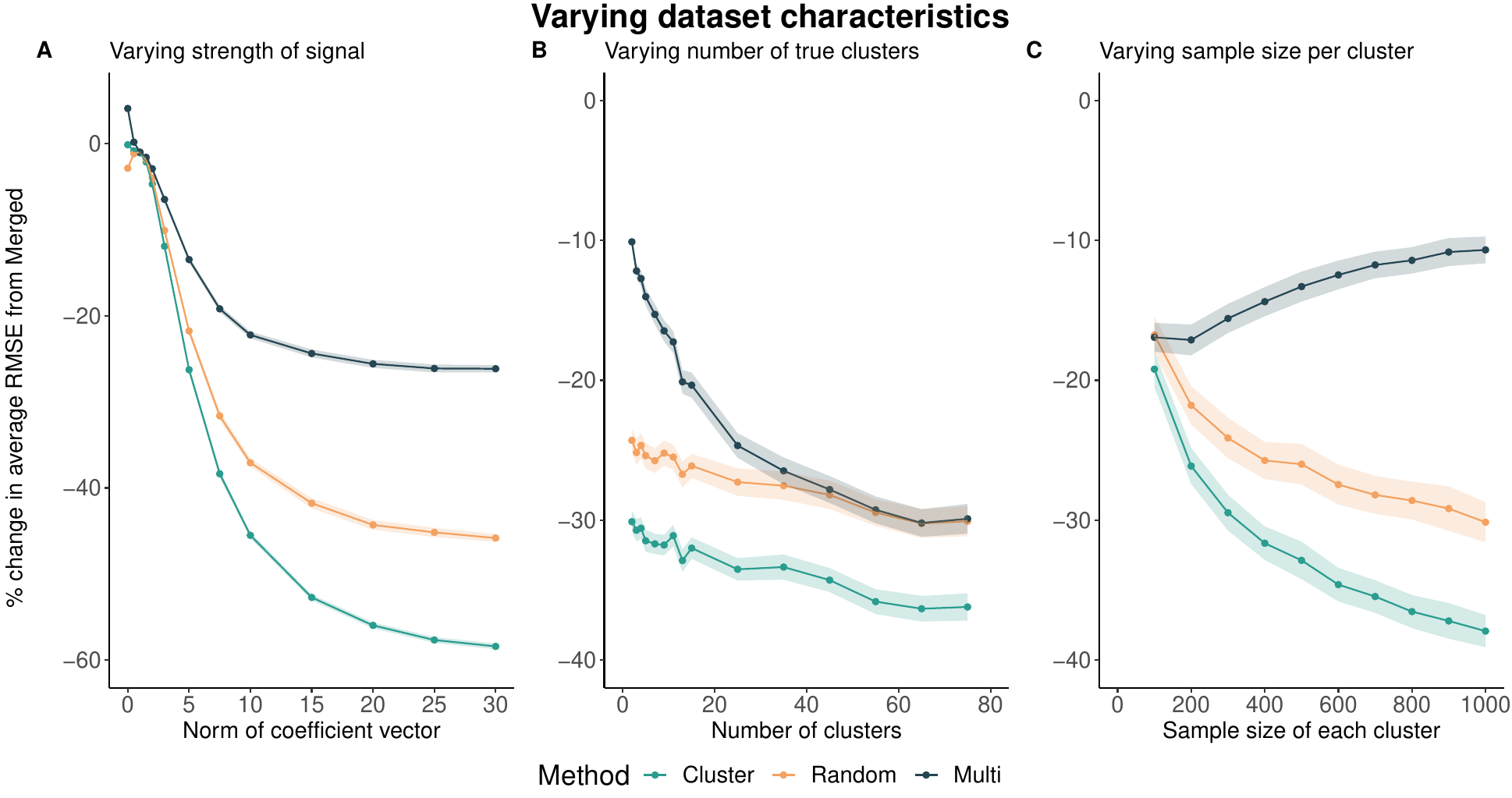}
  \caption{\it Percent change in average RMSE of ensembling approaches (color labeled) compared to the Merged across different data-generating scenarios. All simulations used a quadratic outcome and the non-gaussian cluster generating algorithm (using the `monte' function in the {\tt fungible} package). \textbf{(a)} Varying the magnitude of the coefficients in the outcome-generating model to examine the effect of signal strength on prediction accuracy gains. \textbf{b} Varying the number of true clusters within the training set, while keeping the total sample size constant at 2500. \textbf{(c)} Varying the sample size per cluster, while keeping the total number of clusters per dataset constant at~5.}
  \label{fig:Fig3}
\end{figure*}

\par We next evaluate method robustness across varying dataset characteristics, keeping $k$ at the optimal value determined by Silhouette analysis. Using a quadratic outcome model and non-Gaussian cluster generation, we examine how prediction performance is affected by covariate-outcome signal strength (by varying coefficient norms), the number of true clusters in the training set, and overall sample size.

\par Figure \ref{fig:Fig3}a shows that as covariate signal strength increases (achieved by increasing the norm of the coefficient norm in the simulated outcome model), all ensembling methods improve over \textit{Merged}, with performance gains plateauing at 20-60\%. The \textit{Cluster} consistently outperforms others, especially at higher coefficient norms. The \textit{Multi}'s limited improvement suggests that reducing within-cluster variation is more effective than using true clusters for discerning covariate-outcome relationships. Figure \ref{fig:Fig3}b demonstrates that as true cluster count increases (keeping total sample size at 2500), the \textit{Cluster} remains superior, while the \textit{Multi} approaches the \textit{Random}'s performance. This occurs because \textit{Multi}'s forest count increases with the number of true clusters, aligning with previous observations that more ensemble members improve overall accuracy even as individual model complexity decreases. The \textit{Cluster}'s consistent advantage highlights its benefits regardless of the true data composition. Figure \ref{fig:Fig3}c shows that as sample size increases from 500 to 5000, the performance gap between the \textit{Cluster} and others widens. While the \textit{Multi} becomes less distinguishable from the \textit{Merged} (as the \textit{Merged} accuracy increases proportionally with sample size), the \textit{Cluster} and \textit{Random} continue to improve upon the \textit{Merged} since the optimal number of partitions $k$ also increases proportionally with sample size.


\subsection{Importance of Ensemble Weighting Strategy}

\begin{table*}[t!]
\centering
\begin{tabular}{|l|c|c|c|c|c|c|c|c|} 
 \hline
 \textbf{No.\ of Clusters $K$} & 2 & 5 & 10 & 20 & 30 & 50 & 70 &  80 \\
 \hline\hline
 \textbf{Simple Averaging}  & 11.93 & 29.97 & 43.60 & 59.63 & 68.17 & 79.09 & 86.65 & 89.48 \\ 
  \hline
  \textbf{Stacked Regression} & -9.88 & -13.35 & -19.73 & -25.21 & -27.73 & -29.99 & -31.96 & -32.27 \\ 
 \hline
\end{tabular}
\caption{\it Simple averaging vs. stacked regression weights. Percent change in average RMSE compared to Merged for differently-weighted ensembles built on estimated clusters determined by k-means, using the non-gaussian dataset simulation framework. The first row shows results for weighting each forest equally and the second row depicts results from using stacked regression weights. }
\label{table:T1}
\end{table*}

Our next simulation explores the impact of ensemble weighting on prediction performance, focusing on the advantages of cross-cluster stacked regression weights over other approaches. Table \ref{table:T1} compares the percent change in average RMSE relative to the \textit{Merged} for ensembles using simple averaging versus stacked regression weights. As $k$ increases to its optimal value around 80, stacking-based ensembles continuously improve while equal weighting-based ensembles show declining performance. The best performance for equal weighting occurs at $k=2$, at which it is still significantly worse than the \textit{Merged}. This indicates that not all clusters from k-means produce accurate forest predictors, but certain combinations greatly outperform the \textit{Merged}. Multi-cluster stacking up-weights the cluster-specific learners that display the best cross-cluster prediction ability. Strong cross-cluster performance suggests better learning of the true covariate-outcome relationship and greater generalizability, as clusters are designed for maximal distributional separation.

\section{Data Application: Brain Cancer Genomics}

\par To explore classifier performance on real biological data, we apply our methods to data on tumor samples from patients with various types of brain cancer from the Cancer Genome Atlas project (TCGA) \cite{lgg}. We consider two outcomes: a binary variable indicating tumor grade and type (0 for low-grade gliomas, 1 for high-grade glioblastoma) and a continuous variable measuring the  mutational burden, the total number of DNA mutations found in the tumor. When reporting performance, we use cross-entropy loss for the binary outcome and RMSE for the continuous outcome. Two covariate sets are used in separate analyses: clinical data (513 patients, 50 variables, 12 of which are continuous) with missing data imputed using Random Forest via the 'mice' package, and gene expression data for the same patients previously shown by authors to be separable into clusters (513 patients, 100 genes). For the clinical covariates analysis, we employ the {\tt vscc} package in R, which uses {\tt mclust}'s Gaussian mixture approach for simultaneous variable selection and clustering. Because {\tt mclust} handles a single data type, we apply it to the 12 continuous covariates, demonstrating that CCWF can use clustering algorithms beyond k-means. Gene expression data clustering is based on the original authors' cluster analysis. We implement our ensembling strategies as follows: \textit{Merged} remains a single 500-tree forest trained on all 50 covariates. \textit{Subset Merged} refers to a single 500-tree forest trained only on the 12 continuous variables. \textit{Sample Weighted} refers to first clustering training data using {\tt vscc}, training 100-tree forests on each cluster, and ensembling the learners with weights proportional to each cluster's sample size. \textit{Stack Ridge} follows the same clustering and forest-training paradigm as \textit{Sample Weighted}, but instead ensembles learners using stacked regression weights with a ridge constraint. For each iteration, we randomly select 100 samples for testing and use the rest for training, repeating this process 500 times to obtain distributional and median prediction accuracy measures.

\begin{figure*}[t]
  \centering
  \includegraphics[width=1\linewidth]{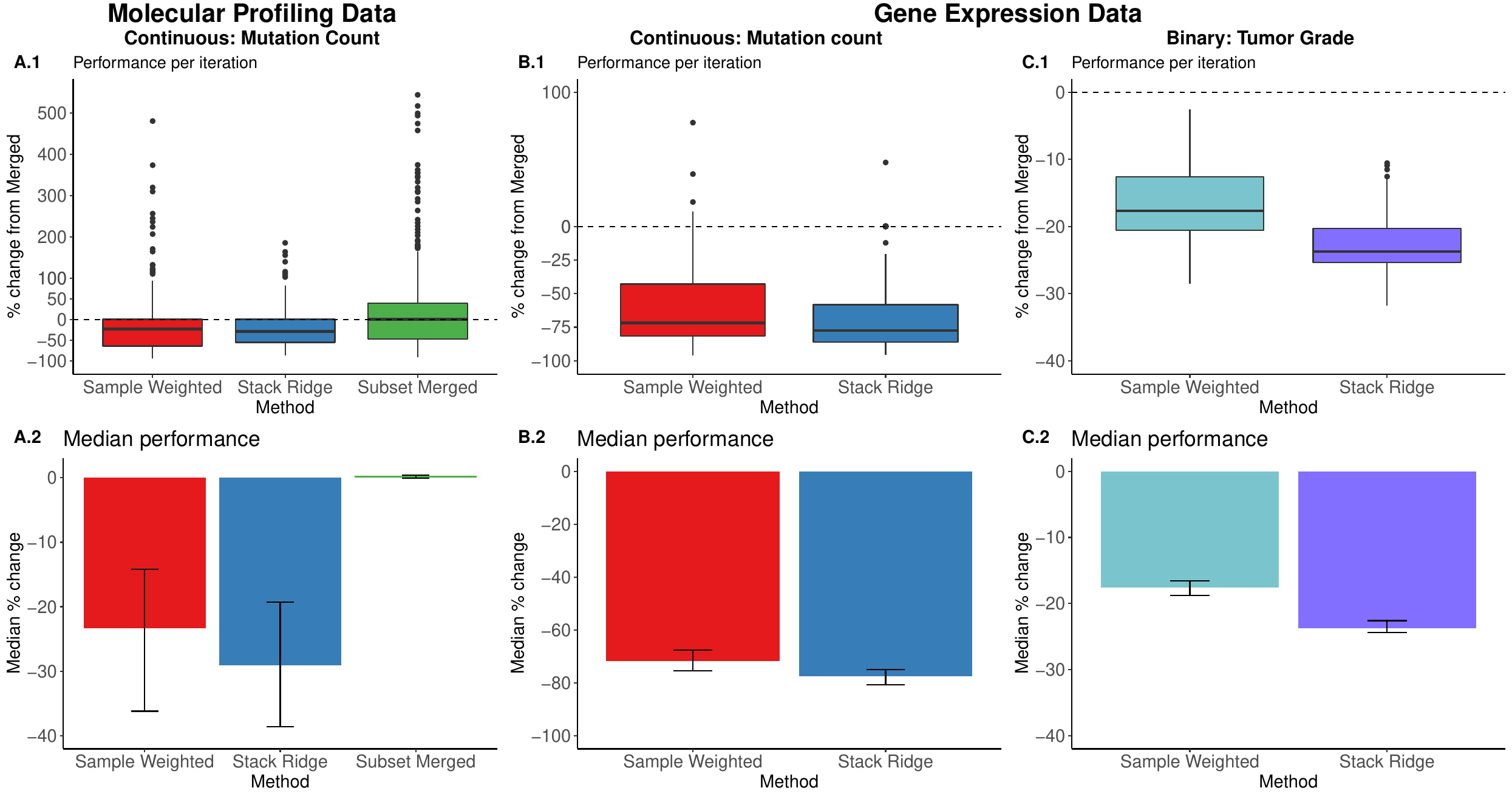}
  \caption{\it Percent change in average performance of ensembling approaches (color labeled) compared to the Merged across two sets of covariate data and two different outcomes from the LGG study. Performance on continuous outcomes in \textbf{(A.1)} and \textbf{(B.1)} is assessed with RMSE, and binary outcome performance in \textbf{C.1} is assessed through cross-entropy loss. and The top row shows 500 iterations of splitting the available data into training and test sets, while the second row displays the median over all iterations with associated confidence intervals. \textbf{(A.1)} Molecular profiling covariate data, mutation count outcome. \textbf{(B.1)} Gene expression covariate data, mutation count outcome. \textbf{(C.1)} Gene expression covariate data, tumor grade outcome. }
  \label{fig:F6}
\end{figure*}

\par Figure \ref{fig:F6} displays the results: we observe substantial improvement of clustering methods over merging approaches across all covariate types and outcomes considered. For molecular profiling data (Fig. \ref{fig:F6}A), ensembles show 20-30\% median improvement over merged-based learners in predicting mutation numbers, indicating {\tt vscc}'s effectiveness for ensemble cluster construction. Gene expression data (Fig. \ref{fig:F6}b) shows larger improvements, with \textit{Sample Weighted} and \textit{Stack Ridge} achieving around 75\% median improvement over the \textit{Merged}. The gene expression clusters used to build the ensemble were previously found to align with biologically relevant characteristics \citep{lgg}. While \textit{Stack Ridge} consistently outperforms \textit{Sample Weighted} across prediction tasks, the difference is modest, suggesting the weighting scheme's influence on prediction ability varies by context. These results demonstrate CCWF's robustness across different outcomes, covariates, and cluster constructions in a real biological setting. Note that in this experiment, we have displayed the top performing approaches; in this case, the \textit{Unweighted} and \textit{Random} approaches are not competitive. 

\section{Conclusions}

\par We introduce Cross-Cluster Weighted Forests (CCWF): ensembles of Random Forests (RFs) trained on estimated clusters with weights rewarding cross-cluster performance. Across theoretical analyses, simulations, and real biological data with feature-distribution heterogeneity, CCWFs yield more generalizable and accurate predictors than standard RFs. Theoretically, we analyze Centered Random Forest learners in a simplified CCWF setup and derive an asymptotic limit of the ensemble-to-merged RMSE ratio, showing that ensembling strictly improves on merging through bias reduction; this limit holds well in our experiments. Our findings highlight the role of data partitioning in RF algorithms. Dividing data by true clusters or studies is not always optimal; reducing within-cluster covariate heterogeneity enhances learning efficiency even at smaller sample sizes. Cluster-based ensembling further outperforms merging when the variables driving cluster formation have low overlap with those most predictive of the outcome. RFs are among the most widely applied machine learning tools, with thousands of published applications across diverse data types. We have identified a simple method that yields substantial improvements, and we believe that the architecture we outlined, as a general concept, may provide further important insight into combining simple and successful machine learning tools via multi-layer approaches. Several limitations remain. First, the theoretical result currently requires uniform clusters, CRFs, and partitions based on known structure; extensions to flexible data distributions, overlapping clusters, and algorithmically-determined cluster partitions remain future work. Second, the scope of our empirical analysis is restricted to biological tabular datasets with moderate dimension and defined heterogeneity. Future work can investigate whether the CCWF approach holds utility for non-biological datasets, which could greatly increase the scope of its utility. Third, our baselines are restricted to RF variants; future work should compare CCWF against other tabular methods such as gradient-boosted trees. 

\paragraph{Software and Reproducibility}
All code to replicate analyses from this paper can be found at \url{https://github.com/m-ramchandran/cross-cluster}. The CCWF method is implemented within the {\tt rCCWF} R package at \url{https://github.com/m-ramchandran/rCCWF} and the {\tt PyCCWF} Python package available through {\tt PyPi}.

\begin{ack}
Maya Ramchandran was supported by NIH-NCI Training Grant T32CA009337 at Harvard T.H. Chan School of Public Health. Rajarshi Mukherjee was partially supported by NSF Grant EAGER-1941419. Giovanni Parmigiani was supported by grant NSF-DMS 1810829.
\end{ack}

\bibliographystyle{plainnat}
\bibliography{staix_2026_sample}

\begin{appendix}

\section{Proofs for Theoretical Analysis}\label{appn} 

\subsection{Centered Random Forests}
\label{sec:centered_forest_defs}
We define a Random Forest as an ensemble of base regression trees
$\{f_n(\bx_{\star};\theta_m, \mathcal{D}n), m \geq 1\}$, where $\theta_m$ are i.i.d. copies of a randomizing variable $\theta$ determining node splits. The forest-level prediction is $\bar{f}_n(\bx_{\star};\theta, \mathcal{D}_n) = \E_{\theta}\left[f_n(\bx_{\star};\theta, \mathcal{D}_n)\right]$, typically estimated by averaging predictions from multiple trees trained on bootstrap samples \citep{breiman2001}. In this section, we analyze the Centered Random Forest method proposed in \cite{Breiman2004}, which simplifies the original algorithm by assuming $\theta$ is independent of the training sample $D_n$. This excludes bootstrapping and data-dependent tree-building strategies but permits $\theta$ to be based on a second sample $\mathcal{D}_n^{'} \stackrel{\rm i.i.d.}{\sim} \mathcal{D}_n$, enabling the splits to be optimized in a manner closely resembling how they would have been with the actual training set. We build upon the analytical work to characterize this model in \cite{klusowski2020} and \cite{biau2012}, keeping our notation consistent whenever possible. 

Centered Random Forest trees are constructed over $\log_2c_n$ iterations, where $c_n \geq 2$ is a user-defined parameter that possibly depends on $n$. Each tree is built as follows: at each node, a coordinate of $\bx$ is selected, with the $j^{th}$ feature having a probability $p_{nj} \in (0,1)$ of being selected, and splits occur at the midpoint of the chosen coordinate. Following from Klusowski (2020), $p_{nj} \to 1/S$ as $n \to \infty$. 
Each tree outputs the average $Y_i$ for $\bx_i$ in the same leaf node partition as $\bx_{\star}$, denoted $A_n(\bx_{\star}, \theta)$. 
The individual tree predictor can therefore be represented as $f_n(\bx_{\star};\theta, \mathcal{D}_n) = \frac{\sum_{i = 1}^n Y_i \mathbbm{1}\{\bx_{\mathbf{i}} \in A_n(\bx_{\star}, \theta)\}}{\sum_{i = 1}^n \mathbbm{1}\{ \bx_{\mathbf{i}} \in A_n(\bx_{\star}, \theta )\}} \mathbbm{1}\{\epsilon_n(\bx_{\star}, \theta)\}$, where $\epsilon_n(\bx_{\star}, \theta )$ is the event that \\
$\sum_{i = 1} \mathbbm{1}\{\bx_{\mathbf{i}} \in A_n(\bx_{\star}, \theta )\} > 0 $; that is, that there is at least one training point that falls within the same leaf node as $\bx_{\star}$. 
We can then obtain the expected prediction made by the Centered Random Forest by taking the expectation of the individual tree predictors with respect to the randomizing variable $\theta$: 
\small
$\bar{f}_n(\bx_{\star};\theta, \mathcal{D}_n) = \sum_{i = 1}^n Y_i \mathbbm{E}_{\theta}\left[\frac{\mathbbm{1}\{\bx_{\mathbf{i}} \in A_n(\bx_{\star}, \theta)\}}{\sum_{i = 1}^n \mathbbm{1}\{\bx_{\mathbf{i}} \in A_n(\bx_{\star}, \theta)\}} \mathbbm{1}\{\epsilon_n(\bx_{\star}, \theta)\} \right]$
\normalsize

\subsection{Statement of Lemma A.1}

\begin{lemma}\label{lemma:forestpredictions}

\begin{enumerate}
    \item [(i)]
For data divided into $k$ clusters, the predictions of the $k$ cluster-level forests within the \textit{Ensemble} learner on new point $\bx_{\star}$ can be expressed as
\begin{align*}
    \hat{Y}_b(\bx_{\star}; \theta, \mathcal{D}_n) &= \sum_{i = 1}^n Y_i\mathbbm{E}_{\theta}\left[ W_{ib}(\bx_{\star}, \theta)\right]
\end{align*}
where
{\scriptsize
\begin{align*}
    W_{ib}(\bx_{\star}, \theta) &= \frac{\mathbbm{1}_{\{\bx_{\mathbf{i}} \in A_n(\bx_{\star}, \theta)\}} \mathbbm{1}_{\{\bx_i \in [\frac{b-1}{k}, \frac{b}{k}]^S\}} \mathbbm{1}_{\{\bx_{\star} \in [\frac{b-1}{k}, \frac{b}{k}]^S\}}}{N_b(\bx_{\star}, \theta)} \\
    & \times \mathbbm{1}_{\{\epsilon_{nb}(\bx_{\star}, \theta)\}}\\
    N_b(\bx_{\star}, \theta) &= \sum_{i = 1}^n \mathbbm{1}_{\{\bx_{\mathbf{i}} \in A_n(\bx_{\star}, \theta)\}} \mathbbm{1}_{\bx_i \in [\frac{b-1}{k}, \frac{b}{k}]^S\}} \mathbbm{1}_{\bx_{\star} \in [\frac{b-1}{k}, \frac{b}{k}]^S\}}
\end{align*}
}
and $\epsilon_{nb}(\bx_{\star}, \theta)$ is the event that $N_b(\bx_{\star}, \theta) > 0$ for $b = 1, \hdots, k$; $N_b(\bx_{\star}, \theta)$ represents the number of total training points from $\X_b$ that fall into the same partition of the test point $\bx_{\star}$ given that $\bx_{\star} \in [\frac{b-1}{k}, \frac{b}{k}]^S$

Specifically, the predictions of the overall \textit{Ensemble} as 
\begin{align*}
    \hat{Y}_E(\bx_{\star} ;\theta, \mathcal{D}_n) &= \sum_{i = 1}^n \sum_{b = 1}^k Y_i\mathbbm{E}_{\theta}\left[ W_{ib}(\bx_{\star}, \theta)\right] \\
    &=  \sum_{i = 1}^n Y_i \mathbbm{E}_{\theta}\left[ W_{i}(\bx_{\star},\theta) \right]
\end{align*}

where $W_{i}(\bx_{\star},\theta) = \sum_{b = 1}^k W_{ib}(\bx_{\star}, \theta)$. 

\item [(ii)]
The predictions of the \textit{Merged} learner on new point $\bx_{\star}$ can be expressed as
\begin{align*}
    \hat{Y}_M(\bx_{\star}; \theta, \mathcal{D}_n) &= \sum_{i = 1}^n Y_i\mathbbm{E}_{\theta}\left[ H_{i}(\bx_{\star}, \theta)\right] 
\end{align*}
where $H_{i}(X, \theta) = \frac{\mathbbm{1}_{\{\bx_{\mathbf{i}} \in A_n(\bx_{\star}, \theta) \}}}{N_n(\bx_{\star}, \theta)} \mathbbm{1}_{\left\{\epsilon_n(\bx_{\star}, \theta) \right\}}$ and \\
$N_n(\bx_{\star}, \theta) = \sum_{i = 1}^n \mathbbm{1}_{\{\bx_{\mathbf{i}} \in A_n(\bx_{\star}, \theta) \}}$, the number of total training samples falling into the same box as $\bx_{\star}$. Finally, $\epsilon_n(\bx_{\star}, \theta)$ is the event that $N_n(\bx_{\star}, \theta) > 0$.
\end{enumerate}
\end{lemma}

\subsection{Proof of Theorem \ref{theorem: rf_uniform}}
In this section, we provide a a derivation for the limit of the ratio between the bias of the \textit{Ensemble} and the bias of the \textit{Merged} when the number of clusters in the training and testing data is equal to $k$ for $k \geq 2$. 

Throughout, we use the following result regarding the leaf node containing test point $\bx_{\star}$.  
\begin{remark}
\label{rem: unif_leafsize}
Let $a_{nj}(\bx_{\star}, \theta)$ and $b_{nj}(\bx_{\star}, \theta)$ be the left and right endpoints of $A_{nj}(\bx_{\star}, \theta)$, the j$^{th}$ side of the box containing $\bx_{\star}$, for $j = 1, \hdots, S$. For ease of notation, we will henceforth refer to $a_{nj}(\bx_{\star}, \theta)$ as $a_{nj}$ and $b_{nj}(\bx_{\star}, \theta)$ as $b_{nj}$. $c_{nj}(\bx_{\star}, \theta)$ represents the number of times that the j$^{th}$ coordinate is split, with the total number of splits across all coordinates set to equal $\log_2{c_n}$ for some constant $c_n > 2$. For ease of notation, we will  suppress the dependencies of $a_{nj}, b_{nj}$, and $c_{nj}$ on $(\bx_{\star}, \theta)$. 
We then observe that each endpoint of $A_{nj}(\bx_{\star}, \theta)$ is a randomly stopped binary expansion of $\bx_{\star}^{(j)}$: 
\begin{align*}
    a_{nj} &\overset{D}{=} \frac{1}{k} \sum_{l = 1}^{c_{nj}} B_{kj} 2^{-l}  + T \\
    b_{nj} &\overset{D}{=} \frac{1}{k}\sum_{l = 1}^{c_{nj}} B_{kj} 2^{-l} + 2^{-{ c_{nj}}} + T
\end{align*}
 No matter the value of T, the length of the j$^{th}$ side of the box is given by:
\begin{align*}
   {\lambda(A_{nj}) = b_{nj} - a_{nj} = \frac{1}{k} 2^{-c_{nj}}}
\end{align*}
The division by the number of clusters $k$ reflects the fact that the test point falls within a cluster of width $\frac{1}{k}^{S}$, so we divide the typical uniform [0,1] binary expansion by the same factor. 
The measure of the box $A_n$ is therefore equal to
\begin{align*}
    \lambda(A_n) &= \prod_{j =1}^{S} \lambda(A_{nj}) \\
    &= \left(\frac{1}{k}\right)^{S} 2^{-\lceil \log_2c_n \rceil} 
\end{align*}
since by construction, $\sum_j c_{nj} = \lceil \log_2c_n \rceil$.
\end{remark}

The rest of the proof is structured as follows:
\begin{enumerate}
\item[(i)] We derive the expression for the ratio of the \textit{Ensemble} and \textit{Merged} biases
\item[(ii)] We derive expressions for the terms $W_1 W_2^{'}$ and $H_1 H_2^{'}$
\item[(iii)] We use (ii) to factor the expressions for the \textit{Ensemble} and \textit{Merged} biases into two terms
\item[(iv)] We provide upper and lower bounds for the first term of the factorization
\item[(v)] We use these bounds to bound the overall limiting ratio
\item[(vi] We simplify the second term of the factorization
\item[(vii)] We combine all previous steps and take limits to complete the proof
\end{enumerate}

\begin{proof}
\hfill
\begin{enumerate}
\item[(i)] The leading term of the squared bias for the \textit{Ensemble} can be represented as
\begin{align*}
& n(n-1)\mathbbm{E}_{\bx_{\star}, \mathcal{D}_n}\left[ \mathbbm{E}_{\theta}\left[ W_{1}\right] (f(\bx_1) - f(\bx_{\star}))\mathbbm{E}_{\theta}\left[ W_{2} \right] (f(\bx_2) - f(\bx_{\star}))\right]  \\
=\text{ } &n(n-1) \mathbbm{E}_{\bx_{\star}, \mathcal{D}_n}\left[ \mathbbm{E}_{\theta}\left[ W_{1} \right] \mathbbm{E}_{\theta}\left[ W_{2} \right] (f(\bx_1) - f(\bx_{\star})) (f(\bx_2) - f(\bx_{\star}))\right] \\
= \text{ } &n(n-1) \mathbbm{E}_{\bx_{\star}, \mathcal{D}_n, \theta, \theta^{'}}\left[ W_{1} W_{2}^{'} (f(\bx_1) - f(\bx_{\star})) (f(\bx_2) - f(\bx_{\star})) \right]
\end{align*}
where $W_{i}(\bx_{\star},\theta) = \sum_{b = 1}^k W_{ib}(\bx_{\star}, \theta)$,  as defined in Lemma \ref{lemma:forestpredictions} (i) and $\theta^{'}$ is an independent copy of $\theta$; thus, for any quantity indexed with the $'$ notation, we replace $A_n(\bx_{\star}, \theta)$ with $A_n(\bx_{\star}, \theta^{'})$. 

The leading term of the squared bias for the \textit{Merged} is equal to
\begin{align*}
& n(n-1)\mathbbm{E}_{\bx_{\star}, \mathcal{D}_n}\left[ \E_{\theta}\left[ H_{1} \right] (f(\bx_1) - f(\bx_{\star}))\E_{\theta}\left[ H_{2} \right] (f(\bx_2) - f(\bx_{\star}))\right] \\
= \text{ } &n(n-1) \mathbbm{E}_{\bx_{\star}, \mathcal{D}_n, \theta, \theta^{'}}\left[ H_{1} H_{2}^{'} (f(\bx_1) - f(\bx_{\star})) (f(\bx_2) - f(\bx_{\star})) \right]
\end{align*}
where $H_1$ and $H_2^{'}$ are defined in Lemma \ref{lemma:forestpredictions} (ii).

Therefore, the limiting ratio of the \textit{Ensemble} bias vs. the \textit{Merged} bias is 

\begin{align*}
\lim_{n \to \infty} \sqrt{\frac{\mathbbm{E}_{\bx_{\star}, \mathcal{D}_n, \theta, \theta^{'}}\left[ W_{1} W_{2}^{'} (f(\bx_1) - f(\bx_{\star})) (f(\bx_2) - f(\bx_{\star}))\right]}{\mathbbm{E}_{\bx_{\star}, \mathcal{D}_n, \theta, \theta^{'}}\left[H_{1} H_{2}^{'} (f(\bx_1) - f(\bx_{\star}))(f(\bx_2) - f(\bx_{\star}))\right]}}
\end{align*}

For the rest of the proof, we focus on evaluating the ratio within the square root, and applying the square root function as the last step. 

\item[(ii)] To simplify this further, we expand the quantities $W_{1} W_{2}^{'}$ and $H_{1} H_{2}^{'}$. 

\begin{align*}
    \ & W_{1} W_{2}^{'}\\ &= \mathbbm{1}_{\{\bx_{\mathbf{1}} \in A_n\}} \mathbbm{1}_{\{\bx_{\mathbf{2}} \in A_n^{'}\}} \\ 
    &\times \left[ \sum_{b = 1}^k \frac{\mathbbm{1}_{\{\bx_1 \in [\frac{b-1}{k}, \frac{b}{k}]^S\}} \mathbbm{1}_{\{\bx_{\star} \in [\frac{b-1}{k}, \frac{b}{k}]^S\}}}{N_b} \mathbbm{1}_{\{\epsilon_{nb}\}} \right] 
    \times \left[ \sum_{b = 1}^k \frac{\mathbbm{1}_{\{\bx_1 \in [\frac{b-1}{k}, \frac{b}{k}]^S\}} \mathbbm{1}_{\{\bx_{\star} \in [\frac{b-1}{k}, \frac{b}{k}]^S\}}}{{N_b}^{'}} \mathbbm{1}_{\{\epsilon_{nb}^{'}\}} \right] \\
    &= \mathbbm{1}_{\{\bx_{\mathbf{1}} \in A_n\}} \mathbbm{1}_{\{\bx_{\mathbf{2}} \in A_n^{'}\}} \\ 
    &\times\left[\sum_{b = 1}^k \frac{\mathbbm{1}_{\{\bx_1 \in [\frac{b-1}{k}, \frac{b}{k}]^S\}}\mathbbm{1}_{\{\bx_2 \in [\frac{b-1}{k}, \frac{b}{k}]^S\}}[\mathbbm{1}_{\{\bx_{\star} \in [\frac{b-1}{k}, \frac{b}{k}]^S\}}]^2 \mathbbm{1}_{\{\epsilon_{nb}\}} \mathbbm{1}_{\{\epsilon_{nb}^{'}\}}}{N_b N_b^{'}}
     \right]
\end{align*}
The cross terms of the product are equal to 0, since $\bx_{\mathbf{1}}, \bx_{\mathbf{1}}, \text{ and } \bx_{\mathbf{\star}}$ have to fall in the same cluster for any term to be non-zero.

Similarly, 
\begin{align*}
    H_{1} H_{2}^{'} &= \mathbbm{1}_{\{\bx_{\mathbf{1}} \in A_n\}} \mathbbm{1}_{\{\bx_{\mathbf{2}} \in A_n^{'}\}} \times \frac{\mathbbm{1}_{\left\{\epsilon_n \right\}}}{N_n} \times \frac{\mathbbm{1}_{\left\{\epsilon_n^{'} \right\}}}{N_n^{'}} 
\end{align*}
\item[(iii)]
We now simplify these quantities using the following definitions and assumptions: 
Define 
\begin{align*}
    T_b &= \sum_{i \geq 3} \mathbbm{1}_{\{\bx_i  \in A_n\}}\mathbbm{1}_{\{\bx_i  \in [\frac{b-1}{k}, \frac{b}{k}]^S\}} \mathbbm{1}_{\{\bx_{\star} \in [\frac{b-1}{k},  \frac{b}{k}]^S\}} 
\end{align*}
for $b = 1, \hdots, k$
and let $T_b^{'}$ be the equivalent expressions based on $\theta^{'}$. We make the assumption that the quantities $\mathbbm{1}_{\{\epsilon_{nb}\}}$, $b = 1, \hdots, k$, as well as $\mathbbm{1}_{\{\epsilon_{n}\}}$, are all non-zero; that is, that there is always at least one training point that falls within $A_n(\bx_{\star}, \theta)$,  within the box (leaf node) containing the test point, regardless of the test point's true cluster assignment. Without loss of generality, we can therefore order the dataset so that one of $\bx_1$, $\bx_2$ falls within $A_n(\bx_{\star}, \theta)$ and the other falls in another cluster. For any test point, we can identify at least $k$ training data points that belong to each distinct cluster out of which one is guaranteed to fall within the same leaf node as the test point. 
There are therefore a minimum of $k/2$ different arrangements of the training set that meet this criteria for all $b = 1, \hdots, k$.

For test points belonging to cluster $b$, we denote the event that this assumption is met by  \\
$d_b = \mathbbm{1}_{\left\{\epsilon_{nb} >0, \sum_{i = 1}^2 \mathbbm{1}_{\{\bx_i \in A_n\}} = 1, \sum_{i = 1}^2 \mathbbm{1}\left\{\bx_i \in \left[\frac{b-1}{k},  \frac{b}{k}\right]^S\right\} = 1\right\}}$.
For any test point belonging to an arbitrary cluster, we denote the event that this assumption is met by $d_n = \mathbbm{1}_{\{\epsilon_{n} >0, \sum_{i = 1}^2 \mathbbm{1}\{\bx_i \in A_n \} = 1\}}$. This event is in fact exactly equivalent to our overall assumption, so by definition, $P(\mathcal{D}_n) = 1$. Therefore, 
\begin{align*}
\E\left[\frac{\mathbbm{1}_{\{\epsilon_{nb}\}}\mathbbm{1}_{\{\epsilon_{nb}^{'}\}}}{N_b N_b^{'}} \bigg| \bx_{\star}\right] &= \E\left[\frac{1}{(1 + T_b)(1 + T_b^{'})} \bigg| \bx_{\star}, d_b \right] \\
&= \frac{\E\left[\frac{1}{(1 + T_b)(1 + T_b^{'})} \bigg| \bx_{\star} \right]}{P(d_b)} \\
&= \frac{k}{2}\E\left[\frac{1}{(1 + T_b)(1 + T_b^{'})} \bigg| \bx_{\star} \right]
\end{align*}
for $b = 1, \hdots, k$.  For the \textit{Merged} learner, we can similarly define 
\begin{align*}
    U &= \sum_{i \geq 3} \mathbbm{1}_{\{\bx_i  \in A_n\}}
\end{align*}
and  
\begin{align*}
\E\left[\frac{\mathbbm{1}_{\{\epsilon_{n}\}}\mathbbm{1}_{\{\epsilon_{n}^{'}\}}}{N_n N_n^{'}} \bigg| \bx_{\star}\right] &= \E\left[\frac{1}{(1 + U)(1 + U^{'})} \bigg| \bx_{\star}, d_n \right] \\
&= \E\left[\frac{1}{(1 + U)(1 + U^{'})} \bigg| \bx_{\star} \right] \\
\end{align*}

Given these definitions, we can further simplify the ratio of the squared biases:  
\begin{align*}
&\frac{\mathbbm{E}_{\bx_{\star}, \mathcal{D}_n, \theta, \theta^{'}}\left[ W_{1} W_{2}^{'} (f(\bx_1) - f(\bx_{\star})) (f(\bx_2) - f(\bx_{\star}))\right]}{\mathbbm{E}_{\bx_{\star}, \mathcal{D}_n, \theta, \theta^{'}}\left[H_{1} H_{2}^{'} (f(\bx_1) - f(\bx_{\star}))(f(\bx_2) - f(\bx_{\star}))\right]} \\
\end{align*}

\item[(iiia)]
The numerator (corresponding to the leading term of the squared bias of the \textit{Ensemble}) simplifies to: 
\begin{align*}
&\mathbbm{E}_{\bx_{\star}, \mathcal{D}_n, \theta, \theta^{'}}\left[ W_{1} W_{2}^{'} (f(\bx_1) - f(\bx_{\star})) (f(\bx_2) - f(\bx_{\star}))\right] \\
= \text{ } & \mathbbm{E}_{\bx_{\star}, \mathcal{D}_n, \theta, \theta^{'}}\Bigg[ \mathbbm{1}_{\{\bx_{\mathbf{1}} \in A_n\}} \mathbbm{1}_{\{\bx_{\mathbf{2}} \in A_n^{'}\}} \times \left[\sum_{b = 1}^k \frac{\mathbbm{1}_{\{\bx_1 \in [\frac{b-1}{k}, \frac{b}{k}]^S\}}\mathbbm{1}_{\{\bx_2 \in [\frac{b-1}{k}, \frac{b}{k}]^S\}}[\mathbbm{1}_{\{\bx_{\star} \in [\frac{b-1}{k}, \frac{b}{k}]^S\}}]^2 \mathbbm{1}_{\{\epsilon_{nb}\}} \mathbbm{1}_{\{\epsilon_{nb}^{'}\}}}{N_b N_b^{'}}\right] \\
= \text{ } & \mathbbm{E}_{\bx_{\star}, \mathcal{D}_n, \theta, \theta^{'}}\Bigg[ \mathbbm{E}\left[\mathbbm{1}_{\{\bx_{\mathbf{1}} \in A_n\}} \mathbbm{1}_{\{\bx_{\mathbf{2}} \in A_n^{'}\}} \times k\left[\frac{\mathbbm{1}_{\{\bx_1 \in [0, \frac{1}{k}]^S\}}\mathbbm{1}_{\{\bx_2 \in [0, \frac{1}{k}]^S\}}[\mathbbm{1}_{\{\bx_{\star} \in [0, \frac{1}{k}]^S\}}]^2 \mathbbm{1}_{\{\epsilon_{n1}\}} \mathbbm{1}_{\{\epsilon_{n1}^{'}\}}}{N_1 N_1^{'}}\right] \bigg| \bx_{\star}, \theta, \theta^{'} \right] \Bigg] \\
= \text{ } & k \times \mathbbm{E}_{\bx_{\star}, \mathcal{D}_n}\Bigg[ \mathbbm{E}\left[\frac{1}{(1 + T_1)(1 + T_1^{'})} \bigg| \bx_{\star}, \theta, \theta^{'}, d_b \right] \\
& \times \mathbbm{E}_{\bx_1, \bx_2}\bigg[\mathbbm{1}_{\{\bx_{\mathbf{1}} \in A_n\}} \mathbbm{1}_{\{\bx_{\mathbf{2}} \in A_n^{'}\}}\mathbbm{1}_{\{\bx_1 \in [0, \frac{1}{k}]^S\}}\mathbbm{1}_{\{\bx_2 \in [0, \frac{1}{k}]^S\}}[\mathbbm{1}_{\{\bx_{\star} \in [0, \frac{1}{k}]^S\}}]^2 \\
& \times (f(\bx_1) - f(\bx_{\star})) (f(\bx_2) - f(\bx_{\star})) \bigg| \bx_{\star}, \theta, \theta^{'}  \bigg] \Bigg]\\
= \text{ } & \frac{k^2}{2} \mathbbm{E}_{\bx_{\star}, \mathcal{D}_n}\Bigg[ \mathbbm{E}\left[\frac{1}{(1 + T_1)(1 + T_1^{'})} \bigg| \bx_{\star}, \theta, \theta^{'} \right] \\
& \times \left(\mathbbm{E}_{\bx_1}\bigg[\mathbbm{1}_{\{\bx_{\mathbf{1}} \in A_n\}} \mathbbm{1}_{\{\bx_1 \in [0, \frac{1}{k}]^S\}}\mathbbm{1}_{\{\bx_{\star} \in [0, \frac{1}{k}]^S\}} (f(\bx_1) - f(\bx_{\star})) \bigg| \bx_{\star}\Bigg]\right)^2 \Bigg]\\
\end{align*}

Various independence relations are used to simplify the expression: $\bx_1 \perp \bx_2$, $\theta \perp \theta^{'}$, and  $T_1 \perp \bx_1, \bx_2$. 

\item[(iiib)]

The denominator (corresponding to the leading term of the squared bias of the \textit{Merged}) simplifies to: 

\begin{align*}
&\mathbbm{E}_{\bx_{\star}, \mathcal{D}_n, \theta, \theta^{'}}\left[H_{1} H_{2}^{'} (f(\bx_1) - f(\bx_{\star}))(f(\bx_2) - f(\bx_{\star}))\right] \\
\text{ } & \mathbbm{E}_{\bx_{\star}, \mathcal{D}_n, \theta, \theta^{'}}\left[\mathbbm{1}_{\{\bx_{\mathbf{1}} \in A_n\}} \mathbbm{1}_{\{\bx_{\mathbf{2}} \in A_n^{'}\}}  \left(\frac{1}{1 + U}\right)\left(\frac{1}{1 + U^{'}}\right) (f(\bx_1) - f(\bx_{\star})) (f(\bx_2) - f(\bx_{\star}))\right] \\ 
\text{ } & \mathbbm{E}_{\bx_{\star}, \mathcal{D}_n, \theta}\left[\left(\frac{1}{1 + U}\right)\left(\frac{1}{1 + U^{'}}\right) \left(\mathbbm{1}_{\{\bx_{\mathbf{1}} \in A_n\}}  (f(\bx_1) - f(\bx_{\star})) \right)^2\right] \\
= \text{ } & \mathbbm{E}_{\bx_{\star}, \mathcal{D}_n}\Bigg[ \mathbbm{E}\left[\frac{1}{(1 + U)(1 + U^{'})} | \bx_{\star}, \theta, \theta^{'} \right] \times \left(\mathbbm{E}_{\bx_1}\left[\mathbbm{1}_{\{\bx_{\mathbf{1}} \in A_n\}} (f(\bx_1) - f(\bx_{\star})) | \bx_{\star}\right]\right)^2 \Bigg]\\
\end{align*}

\item[(iv)]
We now focus on calculating $\mathbbm{E}\left[\frac{1}{(1 + T_1)(1 + T_1^{'})} | \bx_{\star}, \theta, \theta^{'} \right]$ and $\mathbbm{E}\left[\frac{1}{(1 + U)(1 + U^{'})} | \bx_{\star}, \theta, \theta^{'} \right]$. To upper bound these quantities, we use the fact that for $Z \sim \text{Binomial}(m, p)$, $\mathbbm{E}\left[\left(\frac{1}{1+Z} \right)^2\right] \\\leq \frac{1}{(m + 1)(m + 2) p^2}$ along with the Cauchy-Schwarz inequality. To lower bound these quantities, we use the fact that for  $Z \sim \text{Binomial}(m, p)$, one has that $\mathbbm{E}\left[\frac{1}{1+Z}\right] = \frac{1}{(m + 1) p}\left[1-(1-p)^{m+1}\right]$ along with the definition of covariance, which tells us that Cov$(X, Y) = E[XY] - E[X]E[Y] \Rightarrow E[X]E[Y] \leq E[XY]$ for positive Cov$(X, Y)$. 
To move forward, we need to calculate the binomial probabilities associated with $T_1$ and $U$. 

$T_1 \sim \text{Binomial}(n-2, p_e)$, where 
\begin{align*}
    p_e &=\mathbbm{E}\left[\mathbbm{1}_{\{\bx_i  \in A_n\}}\mathbbm{1}_{\{\bx_i  \in [0, \frac{1}{k}]^S\}} \mathbbm{1}_{\{\bx_{\star} \in [0, \frac{1}{k}]^S\}} \right] \\
    &= \mathbbm{E}\left[\mathbbm{1}_{\{\bx_i  \in A_n\}} \bigg| \mathbbm{1}_{\{\bx_i  \in [0, \frac{1}{k}]^S\}} \mathbbm{1}_{\{\bx_{\star} \in [0, \frac{1}{k}]^S\}} \right] \times \mathbbm{E}\left[\mathbbm{1}_{\{\bx_i  \in [0, \frac{1}{k}]^S\}} \right] \times \mathbbm{E}\left[\mathbbm{1}_{\{\bx_{\star} \in [0, \frac{1}{k}]^S\}} \right] \\
    &= \frac{1}{k^2} \prod_{j = 1}^S \frac{b_{nj} - a_{nj}}{\frac{1}{k}} \\
    &= \frac{1}{k^2} 2^{-\lceil \log_2c_n \rceil}
\end{align*}

$U \sim \text{Binomial}(n-2, p_m)$ where 
\begin{align*}
    p_m &= E_{\bx_1} \left[ \mathbbm{1}_{\{\bx_{\mathbf{1}} \in A_n \}} \right] \\
    &= \prod_{j = 1}^S k \times P\left(\mathbbm{1}_{\{\bx_1^{(j)} \in A_{nj}\}} \bigg|\bx_1^{(j)} \in \left[0, \frac{1}{k}\right] \right)\text{P}\left(\bx_1^{(j)} \in \left[0, \frac{1}{k}\right]\right) \\ 
    &= \prod_{j = 1}^S  P\left(\mathbbm{1}_{\{\bx_1^{(j)} \in A_{nj}\}} \bigg|\bx_1^{(j)} \in \left[0, \frac{1}{k} \right], \bx_{\star}^{(j)} \in \left[0, \frac{1}{k} \right] \right) \text{P}\left(\bx_{\star}^{(j)} \in \left[0, \frac{1}{k}\right]\right) \\
    &= \prod_{j = 1}^S \frac{b_{nj} - a_{nj}}{\frac{1}{k}} \times \frac{1}{k} \\
    & = \frac{1}{k} 2^{-\lceil \log_2c_n \rceil}
\end{align*}

Now, we can state that by the definition of covariance, 

\begin{align*}
\mathbbm{E}\left[\frac{1}{(1 + T_1)(1 + T_1^{'})} \bigg| \bx_{\star}, \theta, \theta^{'} \right]  &\geq \mathbbm{E}\left[\frac{1}{(1 + T_1)} \bigg| \bx_{\star}, \theta, \theta^{'} \right] \mathbbm{E}\left[\frac{1}{(1 + T_1^{'})} \bigg| \bx_{\star}, \theta, \theta^{'} \right]\\
&= \left(\mathbbm{E}\left[\frac{1}{(1 + T_1)} \bigg| \bx_{\star}, \theta, \theta^{'} \right] \right)^2  \\
&= \frac{k^4 4^{\lceil \log_2c_n \rceil}}{(n-1)^2} \left[1 - \left(\frac{1}{k^2} 2^{-\lceil \log_2c_n \rceil}\right)^{n-1}\right]^2
\end{align*}

By the Cauchy-Schwarz inequality,
\begin{align*}
    \ & \mathbbm{E}\left[\frac{1}{(1 + T_1)(1 + T_1^{'})} \bigg|\bx_{\star}, \theta, \theta^{'} \right]\\ &\leq \sqrt{\mathbbm{E}\left[\left(\frac{1}{1 + T_1}\right)^2  \bigg|\bx_{\star}, \theta, \theta^{'}\right]} \sqrt{\mathbbm{E}\left[\left(\frac{1}{1 + T_1^{'}}\right)^2  \bigg|\bx_{\star}, \theta, \theta^{'}\right]} \\
    &=\left(\frac{k^2 2^{\lceil \log_2c_n \rceil}}{\sqrt{n(n-1)}}\right)^2 \\
    &= \frac{k^44^{\lceil \log_2c_n \rceil}}{n(n-1)}
\end{align*}

Putting it all together, 
\begin{align*}
 \frac{k^4 4^{\lceil \log_2c_n \rceil}}{(n-1)^2} \left[1 - \left(\frac{1}{k^2} 2^{-\lceil \log_2c_n \rceil}\right)^{n-1}\right]^2 \leq \mathbbm{E}\left[\frac{1}{(1 + T_1)(1 + T_1^{'})} \bigg|\bx_{\star}, \theta, \theta^{'} \right] &\leq \frac{k^44^{\lceil \log_2c_n \rceil}}{n(n-1)}
\end{align*}

Now, we can do a similar calculation for $\mathbbm{E}\left[\frac{1}{(1 + U)(1 + U^{'})} | \bx_{\star}, \theta, \theta^{'} \right]$, yielding:

\begin{align*}
 \frac{k^2 4^{\lceil \log_2c_n \rceil}}{(n-1)^2} \left[1 - \left(\frac{1}{k} 2^{-\lceil \log_2c_n \rceil}\right)^{n-1}\right]^2 \leq \mathbbm{E}\left[\frac{1}{(1 + U)(1 + U^{'})} \bigg|\bx_{\star}, \theta, \theta^{'} \right] &\leq \frac{k^2 4^{\lceil \log_2c_n \rceil}}{n(n-1)}
\end{align*}

\item[(v)] To get an upper bound on our ratio simplified in part (iii), we can take the max of the numerator and the min of the denominator: 

\begin{align*}
&\max \frac{\frac{k^2}{2} \mathbbm{E}_{\bx_{\star}, \mathcal{D}_n}\Bigg[ \begin{array}{c}\mathbbm{E}\left[\frac{1}{(1 + T_1)(1 + T_1^{'})} | \bx_{\star}, \theta, \theta^{'} \right] \\ \times\left(\mathbbm{E}_{\bx_1}\bigg[\mathbbm{1}_{\{\bx_{\mathbf{1}} \in A_n\}} \mathbbm{1}_{\{\bx_1 \in [0, \frac{1}{k}]^S\}}\mathbbm{1}_{\{\bx_{\star} \in [0, \frac{1}{k}]^S\}} (f(\bx_1) - f(\bx_{\star})) \bigg| \bx_{\star}\Bigg]\right)^2  \end{array}\Bigg]}{\mathbbm{E}_{\bx_{\star}, \mathcal{D}_n}\Bigg[ \mathbbm{E}\left[\frac{1}{(1 + U)(1 + U^{'})} | \bx_{\star}, \theta, \theta^{'} \right] \left(\mathbbm{E}_{\bx_1}\left[\mathbbm{1}_{\{\bx_{\mathbf{1}} \in A_n\}} (f(\bx_1) - f(\bx_{\star})) \bigg| \bx_{\star}\right]\right)^2 \Bigg]} \\
&=\frac{\frac{k^2}{2} \mathbbm{E}_{\bx_{\star}, \mathcal{D}_n}\Bigg[ \frac{k^44^{\lceil \log_2c_n \rceil}}{n(n-1)} \left(\mathbbm{E}_{\bx_1}\bigg[\mathbbm{1}_{\{\bx_{\mathbf{1}} \in A_n\}} \mathbbm{1}_{\{\bx_1 \in [0, \frac{1}{k}]^S\}}\mathbbm{1}_{\{\bx_{\star} \in [0, \frac{1}{k}]^S\}} (f(\bx_1) - f(\bx_{\star})) \bigg| \bx_{\star}\Bigg]\right)^2  \Bigg]}{\mathbbm{E}_{\bx_{\star}, \mathcal{D}_n}\Bigg[ \frac{k^24^{\lceil \log_2c_n \rceil}}{(n-1)^2} \left[1 - \left(\frac{1}{k} 2^{-\lceil \log_2c_n \rceil}\right)^{n-1}\right]^2 \left(\mathbbm{E}_{\bx_1}\left[\mathbbm{1}_{\{\bx_{\mathbf{1}} \in A_n\}} (f(\bx_1) - f(\bx_{\star})) \bigg| \bx_{\star}\right]\right)^2 \Bigg]} \\
&= \left(\frac{k^4}{2}\right)\left(\frac{n-1}{n}\right)\left(\frac{1}{\left[1 - \left(\frac{1}{k} 2^{-\lceil \log_2c_n \rceil}\right)^{n-1}\right]^2}\right) \\
&\times \frac{\mathbbm{E}_{\bx_{\star}, \mathcal{D}_n}\Bigg[ \left(\mathbbm{E}_{\bx_1}\bigg[\mathbbm{1}_{\{\bx_{\mathbf{1}} \in A_n\}} \mathbbm{1}_{\{\bx_1 \in [0, \frac{1}{k}]^S\}}\mathbbm{1}_{\{\bx_{\star} \in [0, \frac{1}{k}]^S\}} (f(\bx_1) - f(\bx_{\star})) \bigg| \bx_{\star}\Bigg]\right)^2  \Bigg]}{\mathbbm{E}_{\bx_{\star}, \mathcal{D}_n}\Bigg[ \left(\mathbbm{E}_{\bx_1}\left[\mathbbm{1}_{\{\bx_{\mathbf{1}} \in A_n\}} (f(\bx_1) - f(\bx_{\star})) \bigg| \bx_{\star}\right]\right)^2 \Bigg]}
\end{align*}

We find the lower bound of the ratio by taking the min of the numerator and max of the denominator, yielding: 
\begin{align*}
&\min \frac{\frac{k^2}{2} \mathbbm{E}_{\bx_{\star}, \mathcal{D}_n}\Bigg[\begin{array}{c} \mathbbm{E}\left[\frac{1}{(1 + T_1)(1 + T_1^{'})} | \bx_{\star}, \theta, \theta^{'} \right] \\ \times \left(\mathbbm{E}_{\bx_1}\bigg[\mathbbm{1}_{\{\bx_{\mathbf{1}} \in A_n\}} \mathbbm{1}_{\{\bx_1 \in [0, \frac{1}{k}]^S\}}\mathbbm{1}_{\{\bx_{\star} \in [0, \frac{1}{k}]^S\}} (f(\bx_1) - f(\bx_{\star})) \bigg| \bx_{\star}\Bigg]\right)^2  \end{array}\Bigg]}{\mathbbm{E}_{\bx_{\star}, \mathcal{D}_n}\Bigg[ \mathbbm{E}\left[\frac{1}{(1 + U)(1 + U^{'})} | \bx_{\star}, \theta, \theta^{'} \right] \left(\mathbbm{E}_{\bx_1}\left[\mathbbm{1}_{\{\bx_{\mathbf{1}} \in A_n\}} (f(\bx_1) - f(\bx_{\star})) \bigg| \bx_{\star}\right]\right)^2 \Bigg]} \\
&= \left(\frac{k^4}{2}\right)\left(\frac{n}{n-1}\right) \left[1 - \left(\frac{1}{k^2} 2^{-\lceil \log_2c_n \rceil}\right)^{n-1}\right]^2 \\
&\times \frac{\mathbbm{E}_{\bx_{\star}, \mathcal{D}_n}\Bigg[ \left(\mathbbm{E}_{\bx_1}\Bigg[\mathbbm{1}_{\{\bx_{\mathbf{1}} \in A_n\}} \mathbbm{1}_{\{\bx_1 \in [0, \frac{1}{k}]^S\}}\mathbbm{1}_{\{\bx_{\star} \in [0, \frac{1}{k}]^S\}} (f(\bx_1) - f(\bx_{\star})) \bigg| \bx_{\star}\Bigg]\right)^2  \Bigg]}{\mathbbm{E}_{\bx_{\star}, \mathcal{D}_n}\Bigg[ \left(\mathbbm{E}_{\bx_1}\left[\mathbbm{1}_{\{\bx_{\mathbf{1}} \in A_n\}} (f(\bx_1) - f(\bx_{\star})) \bigg| \bx_{\star}\right]\right)^2 \Bigg]}
\end{align*}

\item[(vi)] We now simplify the quantity 
\begin{align*}
\frac{\mathbbm{E}_{\bx_{\star}, \mathcal{D}_n}\Bigg[ \left(\mathbbm{E}_{\bx_1}\bigg[\mathbbm{1}_{\{\bx_{\mathbf{1}} \in A_n\}} \mathbbm{1}_{\{\bx_1 \in [0, \frac{1}{k}]^S\}}\mathbbm{1}_{\{\bx_{\star} \in [0, \frac{1}{k}]^S\}} (f(\bx_1) - f(\bx_{\star})) \bigg| \bx_{\star}\Bigg]\right)^2  \Bigg]}{\mathbbm{E}_{\bx_{\star}, \mathcal{D}_n}\Bigg[ \left(\mathbbm{E}_{\bx_1}\left[\mathbbm{1}_{\{\bx_{\mathbf{1}} \in A_n\}} (f(\bx_1) - f(\bx_{\star})) \bigg| \bx_{\star}\right]\right)^2 \Bigg]}
\end{align*}

We begin with the numerator, evaluating the expression within the inner expectation: 
\begin{align*}
& \mathbbm{E}_{\bx_1}\Bigg[\mathbbm{1}_{\{\bx_{\mathbf{1}} \in A_n\}} \mathbbm{1}_{\{\bx_1 \in [0, \frac{1}{k}]^S\}}\mathbbm{1}_{\{\bx_{\star} \in [0, \frac{1}{k}]^S\}} f(\bx_1) - f(\bx_{\star}) \bigg| \bx_{\star}\Bigg] \\ 
= \text{ } &\mathbbm{E}_{\bx_1}\Bigg[ f(\bx_1) - f(\bx_{\star})\bigg| \bx_{\mathbf{1}} \in A_n, \bx_1 \in \left[0, \frac{1}{k}\right]^S, \bx_{\star} \in \left[0, \frac{1}{k}\right]^S, \bx_{\star}\Bigg] \\
\times &\mathbbm{E}\left[ \mathbbm{1}_{\{\bx_{\mathbf{1}} \in A_n \}} |  \bx_1 \in \left[0, \frac{1}{k}\right]^S,  \bx_{\star} \in \left[0, \frac{1}{k}\right]^S, \bx_{\star} \right] \times \mathbbm{E}\left[\mathbbm{1}_{\{\bx_1 \in \left[0, \frac{1}{k}\right]^S} \right] \times \mathbbm{E}\left[\mathbbm{1}_{\{\bx_{\star} \in \left[0, \frac{1}{k}\right]^S\}} \right] \\
= \text{ } & \left(\frac{\lambda(A_n)}{k} \right) \mathbbm{E}_{\bx_1}\Bigg[ f(\bx_1) - f(\bx_{\star})\bigg| \bx_{\mathbf{1}} \in A_n, \bx_{\star}\Bigg] 
\end{align*}


Moving onto the denominator, we again evaluate the expression within the inner expectation: 
\begin{align*}
&\mathbbm{E}_{\bx_1}\left[\mathbbm{1}_{\{\bx_{\mathbf{1}} \in A_n\}} (f(\bx_1) - f(\bx_{\star})) \bigg| \bx_{\star}\right] \\
= \text{ } &\mathbbm{E}_{\bx_1}\left[ f(\bx_1) - f(\bx_{\star}) \bigg| \bx_{\mathbf{1}} \in A_n, \bx_{\star}\right] \times \mathbbm{E}\left[ \mathbbm{1}_{\{\bx_{\mathbf{1}} \in A_n \}} \bigg| \bx_{\star} \right] \\
= \text{ } &\mathbbm{E}_{\bx_1}\left[ f(\bx_1) - f(\bx_{\star}) \bigg| \bx_{\mathbf{1}} \in A_n, \bx_{\star}\right] \frac{\lambda(A_n)}{\frac{1}{k}}\\
\end{align*}


Therefore,
\begin{align*}
&\frac{\mathbbm{E}_{\bx_{\star}, \mathcal{D}_n}\Bigg[ \left(\mathbbm{E}_{\bx_1}\bigg[\mathbbm{1}_{\{\bx_{\mathbf{1}} \in A_n\}} \mathbbm{1}_{\{\bx_1 \in [0, \frac{1}{k}]^S\}}\mathbbm{1}_{\{\bx_{\star} \in [0, \frac{1}{k}]^S\}} (f(\bx_1) - f(\bx_{\star})) \bigg| \bx_{\star}\Bigg]\right)^2  \Bigg]}{\mathbbm{E}_{\bx_{\star}, \mathcal{D}_n}\Bigg[ \left(\mathbbm{E}_{\bx_1}\left[\mathbbm{1}_{\{\bx_{\mathbf{1}} \in A_n\}} (f(\bx_1) - f(\bx_{\star})) | \bx_{\star}\right]\right)^2 \Bigg]} \\
= \text{ }& \frac{\left(\frac{\lambda(A_n)}{k} \right)^2 \mathbbm{E}_{\bx_{\star}, \mathcal{D}_n}\Bigg[ \left( \mathbbm{E}_{\bx_1}\Bigg[ f(\bx_1) - f(\bx_{\star})\bigg| \bx_{\mathbf{1}} \in A_n, \bx_{\star}\Bigg] \right)^2  \Bigg]}{\left(k\lambda(A_n) \right)^2 \mathbbm{E}_{\bx_{\star}, \mathcal{D}_n}\Bigg[ \left(\mathbbm{E}_{\bx_1}\left[ f(\bx_1) - f(\bx_{\star}) \bigg| \bx_{\mathbf{1}} \in A_n, \bx_{\star}\right]\right)^2 \Bigg]} \\
= \text{ }& \frac{1}{k^4}
\end{align*}

Note that we do not need to make any assumptions on the form that $f(x)$ takes, since the expressions involving the outcome model cancel between numerator and denominator. 

\item[(vii)]
Putting everything together from parts (v) and (vi), 
\begin{align*}
&\frac{\frac{k^2}{2} \mathbbm{E}_{\bx_{\star}, \mathcal{D}_n}\Bigg[ \begin{array}{c}\mathbbm{E}\left[\frac{1}{(1 + T_1)(1 + T_1^{'})} | \bx_{\star}, \theta, \theta^{'} \right]\\ \times \left(\mathbbm{E}_{\bx_1}\bigg[\mathbbm{1}_{\{\bx_{\mathbf{1}} \in A_n\}} \mathbbm{1}_{\{\bx_1 \in [0, \frac{1}{k}]^S\}}\mathbbm{1}_{\{\bx_{\star} \in [0, \frac{1}{k}]^S\}} (f(\bx_1) - f(\bx_{\star})) \bigg| \bx_{\star}\Bigg]\right)^2  \end{array}\Bigg]}{\mathbbm{E}_{\bx_{\star}, \mathcal{D}_n}\Bigg[ \mathbbm{E}\left[\frac{1}{(1 + U)(1 + U^{'})} | \bx_{\star}, \theta, \theta^{'} \right] \left(\mathbbm{E}_{\bx_1}\left[\mathbbm{1}_{\{\bx_{\mathbf{1}} \in A_n\}} (f(\bx_1) - f(\bx_{\star})) \bigg| \bx_{\star}\right]\right)^2 \Bigg]} \\
\geq \text{ }& \left(\frac{k^4}{2}\right)\left(\frac{n}{n-1}\right)\left[1 - \left(\frac{1}{k^2} 2^{-\lceil \log_2c_n \rceil}\right)^{n-1}\right]^2 \times \frac{1}{k^4} \\
= \text{ }& \left(\frac{n}{n-1}\right) \frac{1}{2} \left[1 - \left(\frac{1}{k^2} 2^{-\lceil \log_2c_n \rceil}\right)^{n-1}\right]^2
\end{align*}

and 

\begin{align*}
&\frac{\frac{k^2}{2} \mathbbm{E}_{\bx_{\star}, \mathcal{D}_n}\Bigg[ \begin{array}{c}\mathbbm{E}\left[\frac{1}{(1 + T_1)(1 + T_1^{'})} | \bx_{\star}, \theta, \theta^{'} \right]\\ \times \left(\mathbbm{E}_{\bx_1}\bigg[\mathbbm{1}_{\{\bx_{\mathbf{1}} \in A_n\}} \mathbbm{1}_{\{\bx_1 \in [0, \frac{1}{k}]^S\}}\mathbbm{1}_{\{\bx_{\star} \in [0, \frac{1}{k}]^S\}} (f(\bx_1) - f(\bx_{\star})) \bigg| \bx_{\star}\Bigg]\right)^2 \end{array} \Bigg]}{\mathbbm{E}_{\bx_{\star}, \mathcal{D}_n}\Bigg[ \mathbbm{E}\left[\frac{1}{(1 + U)(1 + U^{'})} | \bx_{\star}, \theta, \theta^{'} \right] \left(\mathbbm{E}_{\bx_1}\left[\mathbbm{1}_{\{\bx_{\mathbf{1}} \in A_n\}} (f(\bx_1) - f(\bx_{\star})) \bigg| \bx_{\star}\right]\right)^2 \Bigg]} \\
\leq \text{ }& \left(\frac{k^4}{2}\right)\left(\frac{n-1}{n}\right) \times \frac{1}{k^4} \left(\frac{1}{\left[1 - \left(\frac{1}{k} 2^{-\lceil \log_2c_n \rceil}\right)^{n-1}\right]^2}\right) \\
= \text{ }& \left(\frac{n}{n-1}\right) \frac{1}{2} \left(\frac{1}{\left[1 - \left(\frac{1}{k} 2^{-\lceil \log_2c_n \rceil}\right)^{n-1}\right]^2}\right)
\end{align*}

We now take square roots of the minimum and maximum values calculated above since we are interested in the ratio of the \textit{Ensemble} bias vs. the \textit{Merged} bias (not the squared ratio), and take limits as $n \to \infty$. 

\begin{align*}
\lim_{n \to \infty} \frac{\textit{Ensemble bias}}{\textit{Merged bias}} \in 
& \left[\begin{array}{c} \lim_{n \to \infty}\sqrt{\left(\frac{n}{n-1}\right) \frac{1}{2} \left[1 - \left(\frac{1}{k^2} 2^{-\lceil \log_2c_n \rceil}\right)^{n-1}\right]^2},\\ \lim_{n \to \infty}\sqrt{\left(\frac{n}{n-1}\right) \frac{1}{2} \left(\frac{1}{\left[1 - \left(\frac{1}{k} 2^{-\lceil \log_2c_n \rceil}\right)^{n-1}\right]^2}\right)}\end{array}\right] \\
&\to \left[\sqrt{\frac{1}{2}}, \sqrt{\frac{1}{2}}\right] \\
\Rightarrow &\lim_{n \to \infty} \frac{\textit{Ensemble bias}}{\textit{Merged bias}} \to \frac{1}{\sqrt{2}}
\end{align*}
for any $k \geq 2$

\end{enumerate}
\end{proof}



\begin{remark}
\label{rem: biasconvergence}
Convergence of bias terms
\begin{enumerate}
\item[(i)]
From the proof of Theorem 1, we can express the squared bias of the \textit{Ensemble} as:
\begin{align*}
& n(n-1)\mathbbm{E}_{\bx_{\star}, \mathcal{D}_n, \theta, \theta^{'}}\left[ W_{1} W_{2}^{'} (f(\bx_1) - f(\bx_{\star})) (f(\bx_2) - f(\bx_{\star}))\right] \\
 \to \text{ } &\frac{k^{4-2S}}{2}\mathbbm{E}_{\bx_{\star}, \mathcal{D}_n}\Bigg[ \left( \mathbbm{E}_{\bx_1}\Bigg[ f(\bx_1) - f(\bx_{\star})\bigg| \bx_{\mathbf{1}} \in A_n, \bx_{\star}\Bigg] \right)^2\\ 
 \geq \text{ } & O\left(\frac{1}{n} \right)
\end{align*}

The convergence of this term depends on the function $f(x)$, distributions of $\bx_{1} \text{ and } \bx_{\star}$, number of non-sparse variables $S$, the number of clusters $k$, and the total number of splits $c_n$ (which influences the size of the box containing $\bx_{\star}$). Assuming that $\frac{k}{n}$, $\frac{c_n}{n}$, and $\frac{S}{n} \to 0$ as $n \to \infty$, the squared bias converges at a rate greater than $O\left(\frac{1}{n} \right)$
\item[(ii)]
We can express the squared bias of the \textit{Merged} as:
\begin{align*}
& n(n-1)\mathbbm{E}_{\bx_{\star}, \mathcal{D}_n, \theta, \theta^{'}}\left[ H_{1} H_{2}^{'} (f(\bx_1) - f(\bx_{\star})) (f(\bx_2) - f(\bx_{\star}))\right] \\
 \to \text{ } &k^{4-2S}\mathbbm{E}_{\bx_{\star}, \mathcal{D}_n}\Bigg[ \left( \mathbbm{E}_{\bx_1}\Bigg[ f(\bx_1) - f(\bx_{\star})\bigg| \bx_{\mathbf{1}} \in A_n, \bx_{\star}\Bigg] \right)^2\\
  \geq \text{ } & O\left(\frac{1}{n} \right)
\end{align*}
using similar arguments as above. 

\end{enumerate}
\end{remark}

\subsection{Lemma \ref{lemma:variancetozero}}
We now show that the variance of the \textit{Ensemble} and \textit{Merged} asymptotically converge to zero. We assume that $\text{Var}[f(\bx) | \bx] \leq \sigma^2$ for some $\sigma > 0$. 

\begin{lemma}\label{lemma:variancetozero}
\begin{enumerate}
\item[(i)]
Using [Biau 2012, Proposition 2], we can express the variance of the \textit{Ensemble} as 
\begin{align*}
\E[\hat{Y}_E(\bx_{\star}) - \E[\hat{Y}_E(\bx_{\star})|\bx_{\star}]]^2
= \text{ }& n\sigma^2\E\left[\E_{\theta, \theta^{'}}\left[W_1 W_2^{'} \right] \right] \\
= \text{ }& n\sigma^2\frac{k^2}{2} \mathbbm{E}_{\bx_{\star}, \mathcal{D}_n}\Bigg[ \mathbbm{E}\left[\frac{1}{(1 + T_1)(1 + T_1^{'})} \bigg| \bx_{\star}, \theta, \theta^{'} \right] \\ &\times \left(\mathbbm{E}_{\bx_1}\bigg[\mathbbm{1}_{\{\bx_{\mathbf{1}} \in A_n\}} \mathbbm{1}_{\{\bx_1 \in [0, \frac{1}{k}]^S\}}\mathbbm{1}_{\{\bx_{\star} \in [0, \frac{1}{k}]^S\}} \bigg| \bx_{\star}\Bigg]\right)^2 \Bigg]\\
= \text{ }& n\sigma^2\frac{k^2}{2}\frac{k^44^{\lceil \log_2c_n \rceil}}{n(n-1)} \left(\frac{\lambda(A_n)}{\frac{1}{k}} \left(\frac{1}{k} \right)^2 \right)^2 \\
= \text{ }& \frac{\sigma^2 k^{(4-2S)}}{2(n-1)} \\
= \text{ }& O\left(\frac{1}{n} \right)
\end{align*}

using calculations from the proof of Theorem 1, step (iiia), (iv), and (vi) and finite assumptions on $S$ and $k$ outlined in Remark \ref{rem: biasconvergence}. Note that we do NOT need a finite assumption on the number of total splits $c_n$ for the variance to be $O(1/n)$. 

\item[(ii)]
Similarly, we can express the variance of the \textit{Merged} as 
\begin{align*}
&\E[\hat{Y}_M(\bx_{\star}) - \E[\hat{Y}_M(\bx_{\star})|\bx_{\star}]]^2\\
= \text{ }& n\sigma^2\E\left[\E_{\theta, \theta^{'}}\left[H_1 H_2^{'} \right] \right] \\
= \text{ }& n\sigma^2 \mathbbm{E}_{\bx_{\star}, \mathcal{D}_n}\Bigg[ \mathbbm{E}\left[\frac{1}{(1 + U)(1 + U^{'})} \bigg| \bx_{\star}, \theta, \theta^{'} \right] \times \left(\mathbbm{E}_{\bx_1}\bigg[\mathbbm{1}_{\{\bx_{\mathbf{1}} \in A_n\}} \bigg| \bx_{\star}\Bigg]\right)^2 \Bigg]\\
= \text{ }& n\sigma^2\frac{k^24^{\lceil \log_2c_n \rceil}}{n(n-1)} \left(\frac{\lambda(A_n)}{\frac{1}{k}}\right)^2 \\
= \text{ }& \frac{\sigma^2 k^{(3-2S)}}{n-1} \\
= \text{ }& O\left(\frac{1}{n} \right)
\end{align*}

using calculations from the proof of Theorem \ref{theorem: rf_uniform}, step (iiib), (iv), and (vi). 

Therefore, the variance terms converge to 0 as $n 
\to \infty$ and the squared bias asymptotically dominates the MSE. 
\end{enumerate}
\end{lemma}

\section{Simulations Supporting Theoretical Analysis}
\subsection{Simulation study verifying Theorem \ref{theorem: rf_uniform}}
\label{sec:sim_study_theory}
\begin{figure*}[h]
  \centering
  \includegraphics[width=.8\textwidth]{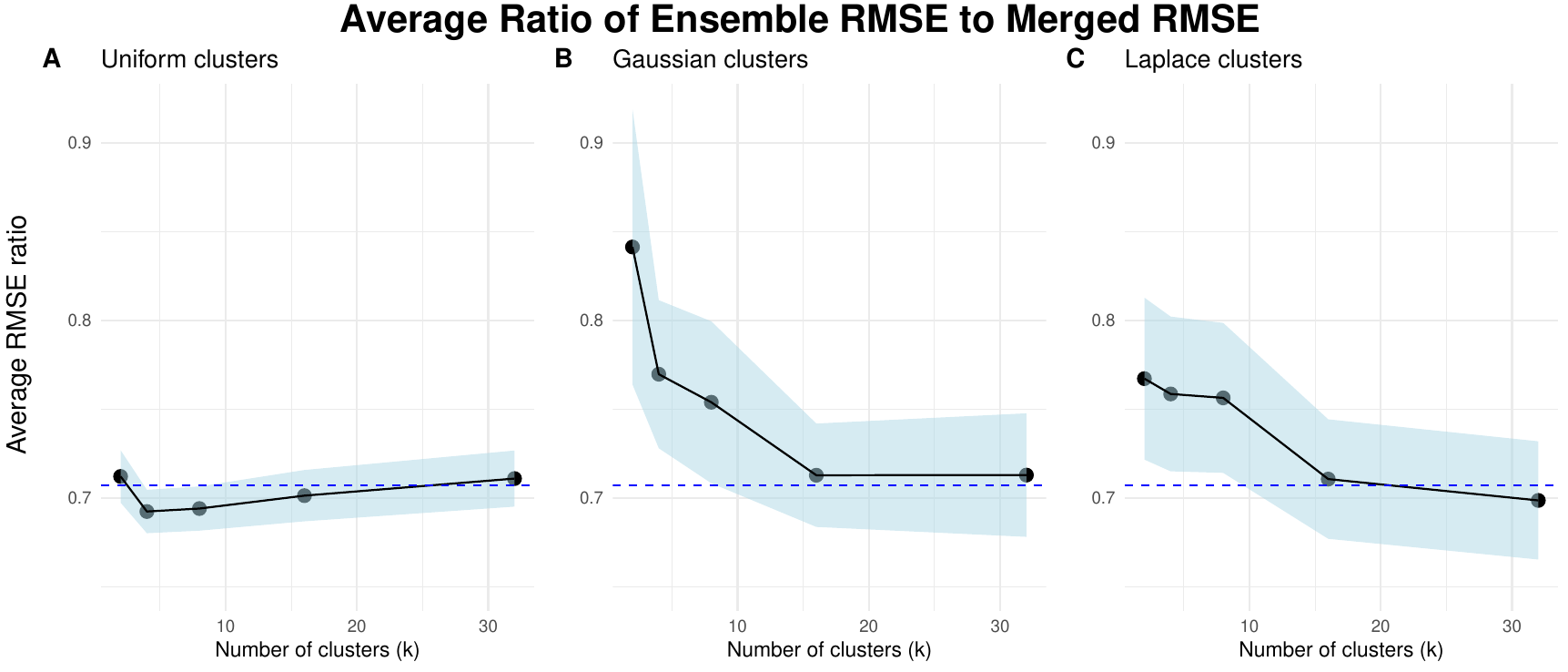}
  \caption{\it Average ratio of \textit{Ensemble} RMSE to \textit{Merged} RMSE as a function of the number of clusters in the training set (n = 5,000). The outcome is simulated as a linear combination of 20 covariates, out of which 10 have zeroed out coefficients to induce sparsity, with added Gaussian noise. The means (points) and 95\% CIs (shaded blue regions) are calculated over 150 replicates at each $k$. The dotted line is at the theoretical limiting ratio $1/\sqrt{2}$. \textbf{(A)} Uniform clusters \textbf{(B)} Multivariate Gaussian clusters \textbf{(C)} Multivariate Laplace clusters}
  \label{fig:F1}
\end{figure*}

\par Figure \ref{fig:F1} presents simulation results illustrating Theorem \ref{theorem: rf_uniform}'s relevance to clustered data generated from three distributional paradigms: uniform, Gaussian, and Laplace. The uniform distribution simulation matches our theoretical setup, with non-overlapping clusters of width $\frac{1}{k}$. In contrast, the Gaussian and Laplace simulations feature overlapping clusters characterized by different location parameters and identity scale parameters. Across all distributions and the number of simulated clusters, the ratio of \textit{Ensemble} to \textit{Merged} RMSEs is consistently close to $\frac{1}{\sqrt{2}}$; as expected, the uniform simulation exhibits a high consistency with the theory. When assumptions are violated, in the Gaussian and Laplace cases, ratios are close to $\frac{1}{\sqrt{2}}$ as $k$ gets large, but can deviate for smaller values. This illustrates that our analytical framework effectively captures the performance improvements of the \textit{Ensemble} across diverse and realistic scenarios at least for large $k$.

\subsection{Examining the Bias and Variance of the \textit{Multi} and \textit{Merged} approaches}
\label{sec:biasvar_multi_merged}

\begin{figure*}[h]
   \centering
   \includegraphics[width=1\linewidth]{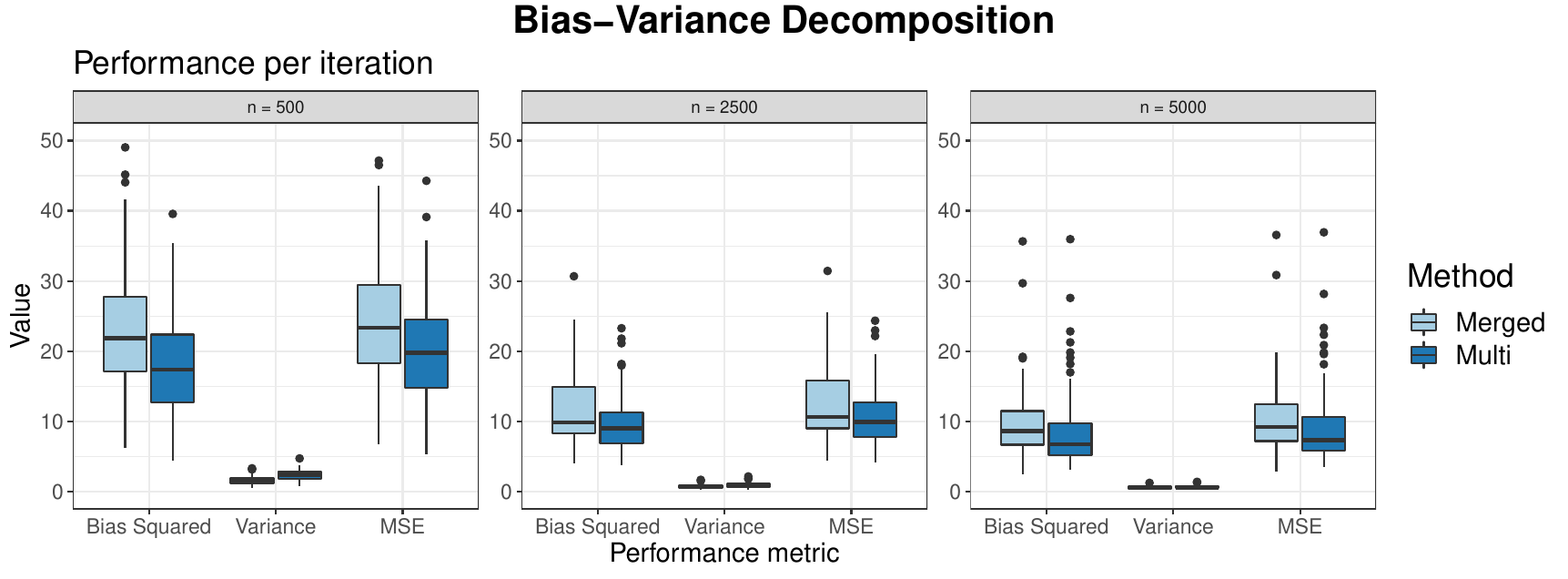}
   \caption{\it Squared bias, variance, and MSE of the Merged and Multi learners (color coded) for datasets ranging from 500-5000 total samples (100-1000 per cluster) and generated using the gaussian cluster framework. Each panel corresponds to a sample size.}
   \label{fig:F5}
 \end{figure*}

\par This simulations decomposes the MSE of \textit{Merged} and \textit{Multi} into their bias and variance terms to individually characterize each using the Gaussian cluster framework outlined in the Simulation setup. This bridges the more simplified setup we explore in the theory with the setup we analyze analytically through simulations and real data examples; instead of non-overlapping uniform clusters, CRFs as the base learners, and equal ensemble weights, we allow for overlapping Gaussian clusters, full RFs for the base learners, and stacking weights for the ensemble. 
As the training sample size increases, the MSE of both ensembling approaches decreases, and across all sample sizes considered, the squared bias term accounts for almost the entirety of the MSE, as expected. There is a far smaller decrease from 2500 to 5000 than from 500 to 2500, suggesting an asymptotic limit to how well the ensembles can perform, and supporting the analytical bounds on the risk from Theorems \ref{theorem: rf_uniform}. All empirical improvements in performance from \textit{Multi} over \textit{Merged} arise through a smaller bias, supporting the validity of analytical results. Furthermore, the average ratio of MSE's between the \textit{Multi} and \textit{Merged} is very similar to the theoretical limiting ratio. We see that in fact the variance is slightly higher for the \textit{Multi} over the \textit{Merged}, but is proportionally so small as to have almost no effect; this directly confirms our theoretical results showing that the variance term converges to 0 faster than the bias term. We limit our comparison in this simulation to the \textit{Multi} rather than full CCWF, since this ensemble construction method can be analytically characterized much more easily than the \textit{Cluster} or \textit{Random}. These results directly show the applicability of our theoretical results to the case in which we use the standard RF algorithm and stacked regression weights.

\section{Additional CCWF Simulations}
\subsection{Robustness to Realistic Dataset Variety}
\label{sec:realistic_dataset_variety}

\begin{figure*}[h]
  \centering
  \includegraphics[width=.8\linewidth]{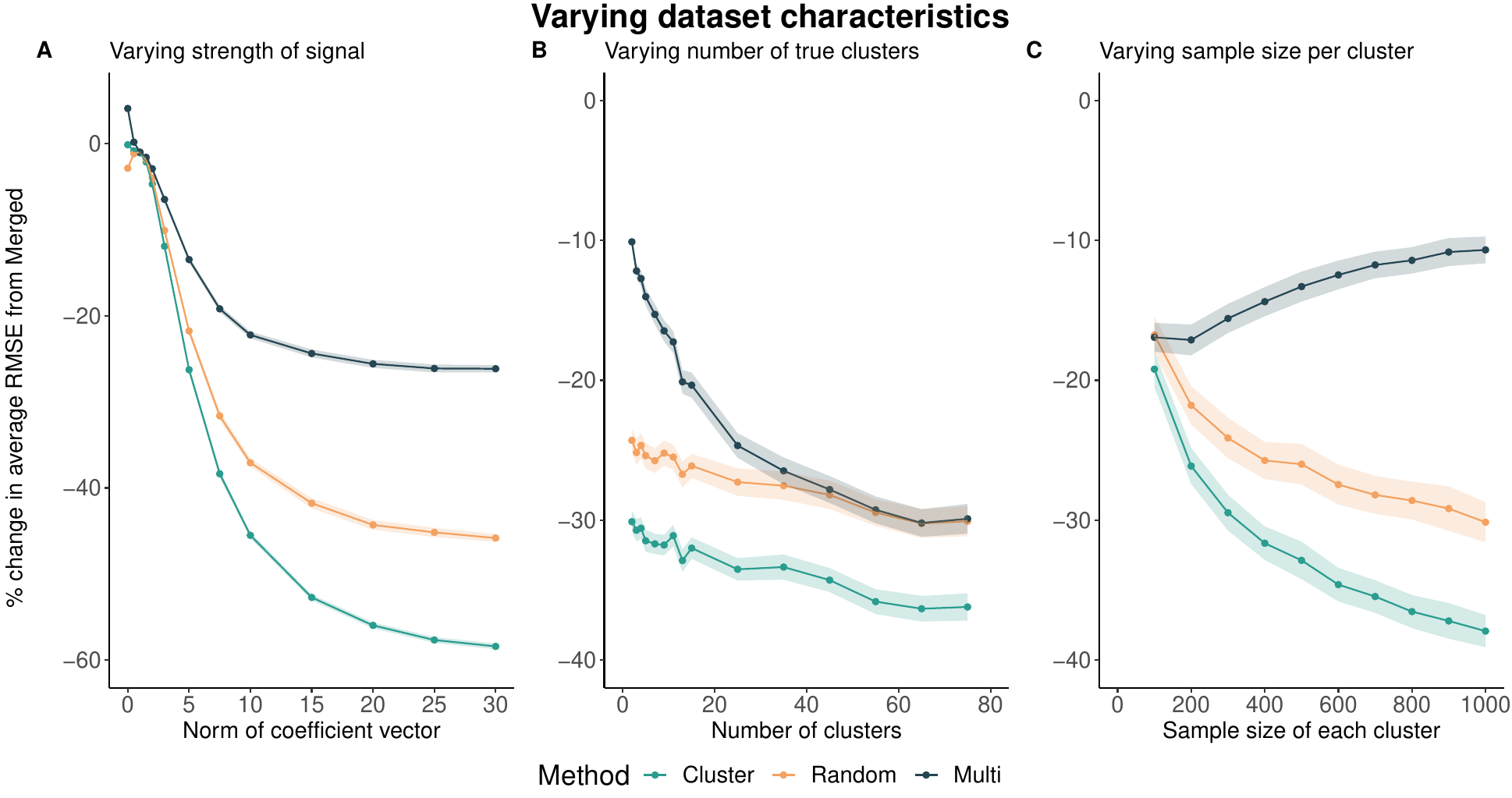}
  \caption{\it Percent change in average RMSE of ensembling approaches (color labeled) compared to the Merged across different data-generating scenarios. All simulations used a quadratic outcome and the non-gaussian cluster generating algorithm (using the `monte' function in the {\tt fungible} package). \textbf{(a)} Varying the magnitude of the coefficients in the outcome-generating model to examine the effect of signal strength on prediction accuracy gains. \textbf{b} Varying the number of true clusters within the training set, while keeping the total sample size constant at 2500. \textbf{(c)} Varying the sample size per cluster, while keeping the total number of clusters per dataset constant at~5.}
  \label{fig:Fig3}
\end{figure*}

\par We next evaluate method robustness across varying dataset characteristics, keeping $k$ at the optimal value determined by Silhouette analysis. Using a quadratic outcome model and non-Gaussian cluster generation, we examine how prediction performance is affected by covariate-outcome signal strength (by varying coefficient norms), the number of true clusters in the training set, and overall sample size.

\par Figure \ref{fig:Fig3}a shows that as covariate signal strength increases (achieved by increasing the norm of the coefficient norm in the simulated outcome model), all ensembling methods improve over \textit{Merged}, with performance gains plateauing at 20-60\%. The \textit{Cluster} consistently outperforms others, especially at higher coefficient norms. The \textit{Multi}'s limited improvement suggests that reducing within-cluster variation is more effective than using true clusters for discerning covariate-outcome relationships. Figure \ref{fig:Fig3}b demonstrates that as true cluster count increases (keeping total sample size at 2500), the \textit{Cluster} remains superior, while the \textit{Multi} approaches the \textit{Random}'s performance. This occurs because \textit{Multi}'s forest count increases with the number of true clusters, aligning with previous observations that more ensemble members improve overall accuracy even as individual model complexity decreases. The \textit{Cluster}'s consistent advantage highlights its benefits regardless of the true data composition. Figure \ref{fig:Fig3}c shows that as sample size increases from 500 to 5000, the performance gap between the \textit{Cluster} and others widens. While the \textit{Multi} becomes less distinguishable from the \textit{Merged} (as the \textit{Merged} accuracy increases proportionally with sample size), the \textit{Cluster} and \textit{Random} continue to improve upon the \textit{Merged} since the optimal number of partitions $k$ also increases proportionally with sample size.

\subsection{CCWF Extension to Multiple Studies}
\label{sec:multistudy}

\begin{figure*}[h]
  \centering
   \includegraphics[width=.65\textwidth]{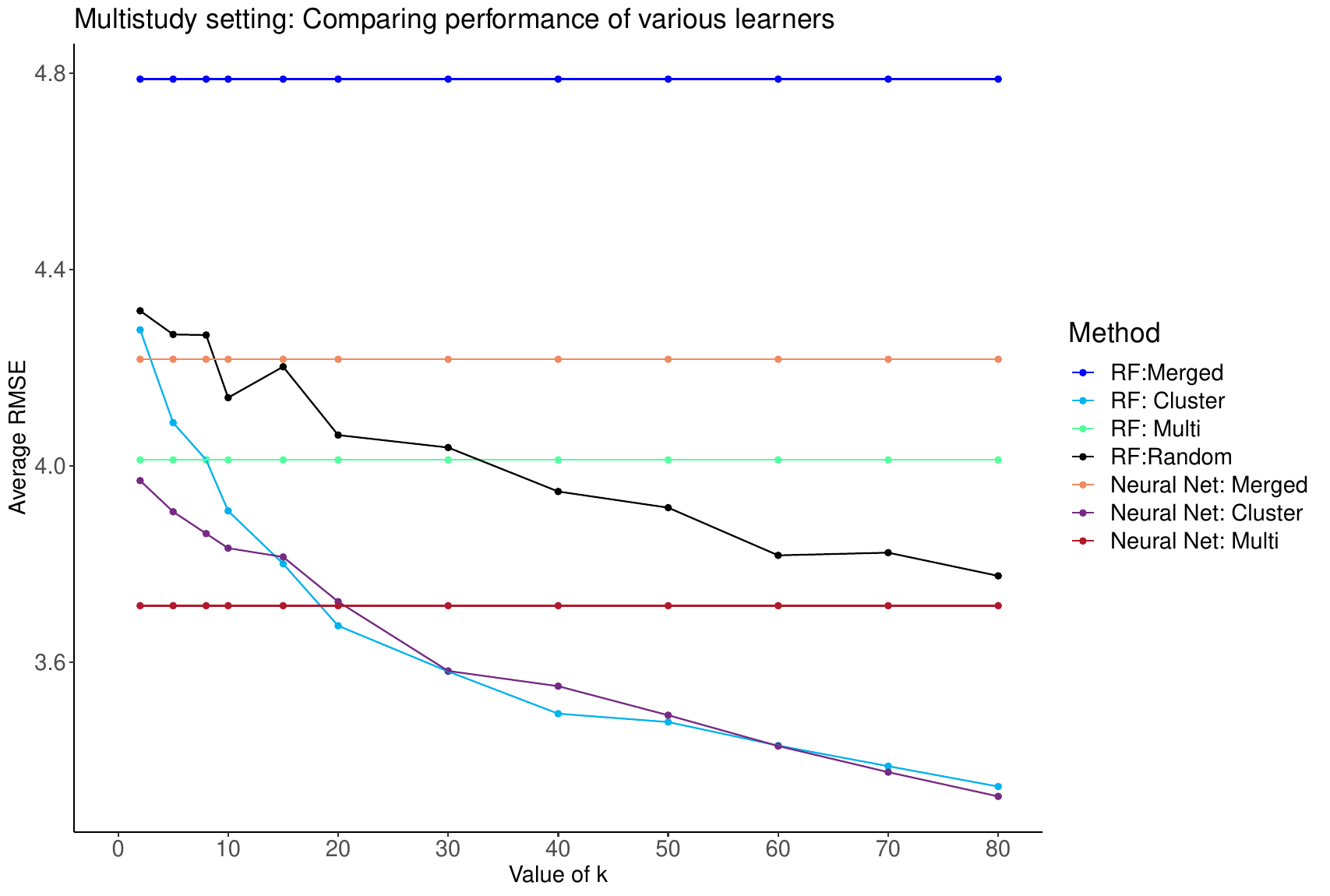}
  \caption{Average RMSE's of ensembling approaches (color labeled) as a function of $k$ in the multi-study analysis. 20 total covariates, 10 associated with the outcome; nonlinear outcome model of the form $y = \beta^T\mathbf{X} + 4.4x_1 - 1.8x_2 + 10\sin(10\pi x_1)$.}
  \label{fig:F7}
\end{figure*}

\par We examine whether we can more optimally partition the total amount of data when multiple studies are available for training that measure the same covariates and outcome variable. The traditional multi-study ensembling paradigm is to train a single learner on each study and combine learners using some weighting strategy such as stacking \citep{Patil2018}. This is analogous to the \textit{Multi} method when we are able to separate data into its true clusters. We now explore whether training k-means on the the merged data (comprising covariate data from all of the training studies) produces improvements comparatively to the single dataset setting. We furthermore evaluate the performance of these approaches on real gene expression data from the CuratedOvarianData repository from Bioconductor in R \citep{Ganzfried2013}. Finally, we explore whether the general strategy of ensembling learners built on clusters also works for Neural Network base learners, and compare the results to ensembles using Random Forest. 

\par The general framework for the simulation setup used to create Figure \ref{fig:F7} is drawn from Ramchandran et. al. (2020) \citep{Ramchandran2020}. We use all 15 studies in CuratedOvarianData that include survival information without any missing data in the features. For N = 250 iterations per value of $k$, we randomly separate the 15 datasets from CuratedOvarianData into 10 training and 5 validation sets. We then generate the outcome using a non-linear model in order to test the ability of the candidate learning methods to detect more difficult covariate-outcome relationships. We simulate baseline levels of coefficient perturbation per study similarly to the setup discussed in Section 4.1. Using either Random Forest or Neural Nets as the base learner, we then construct ensembles using the four main approaches compared throughout this paper. The \textit{Multi} in this case trained a learner on each study to form the final ensemble. All ensembles are built using stacked regression weights with a ridge constraint. 

\par For Random Forest-based ensembles, clustering outperforms study-based partitioning for $k$ greater than the number of training studies. The \textit{Cluster} approach remains most effective, followed by \textit{Random}, \textit{Multi}, and the \textit{Merged}. The true cluster or study-membership does not represent the most effective data partitioning for Random Forest; again, it is more effective to partition the data based on minimizing within-cluster heterogeneity and maximizing across. Overall, these results suggest that for either single or multiple training datasets, the \textit{Cluster} approach should be used for Random Forest learners.

\par For Neural Network-based models, the \textit{Cluster} method still produces improvements over training on true studies or the merged data, though less so than with Random Forest. Notably, the \textit{Cluster} approach yields similar accuracy with both base learning algorithms. While Neural Nets outperform Random Forest using the \textit{Merged} and \textit{Multi} approaches, the \textit{Cluster} method elevates Random Forest-based ensembles to match Neural Net performance. This can be intuitively understood by comparing the \textit{Cluster} strategy to convolutional neural networks, in which the estimated clusters represent the convolutional neighborhoods and the stacking weights delineate the relationship between the tree-learning and the stacking layers. While this analogy does not numerically explain the similar performance of the two algorithms, it provides insight into why we may be seeing these results.

\subsection{Considering a single covariate}
\label{sec:single_covariate}
In this section, we consider datasets with a single covariate, so that we can visualize the differences between the true outcome model and the estimated outcome models from the \textit{Merged} and \textit{Ensemble} as a function of the input covariate. We consider a cubic outcome model and the gaussian cluster-generating framework with two clusters.

\begin{figure*}[!ht]
  \centering
  \includegraphics[width=12cm, height = 7cm]{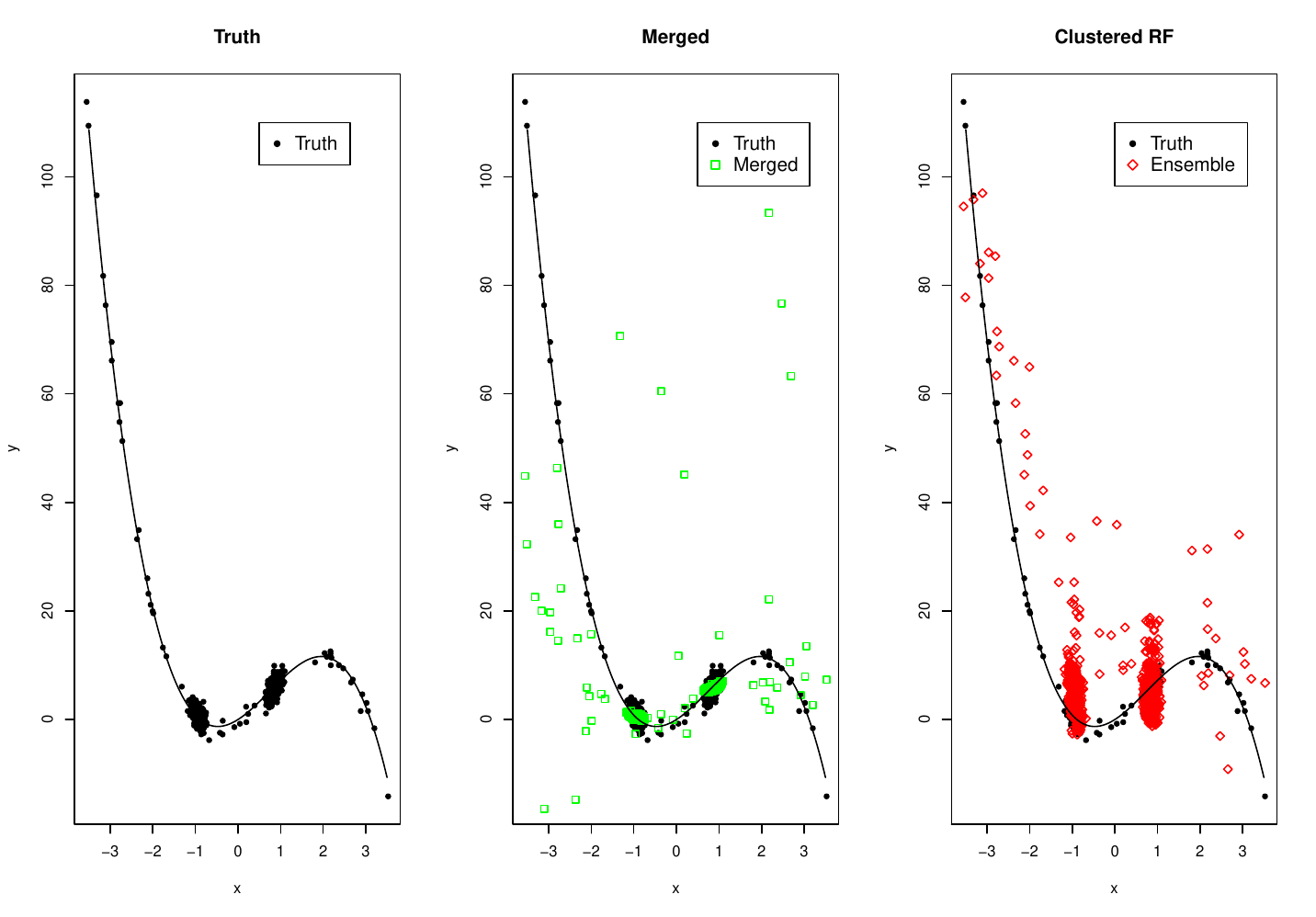}
  \caption{ Cubic outcome model: $y = 5x_1 + 4x_1^2 - 1.8 x_1^3$. The single covariate x is simulated as a mixture model of two gaussian clusters. The leftmost panel displays the covariate data and the true outcomes, as well as the underlying cubic model. The center and right panels displays the predictions of the \textit{Merged} and \textit{Cluster} learners respectively, as well as the underlying true outcome values and model. The \textit{Cluster} approach has an approximately $15\%$ gain in predictive performance over the \textit{Merged}.}
  \label{fig:Sup1}
\end{figure*}

The true model is visualized by the black line in all of the panels and the true outcome values corresponding to the generated covariate data is displayed by the black points. The predictions by the \textit{Merged} and \textit{Ensemble} are depicted in green and red respectively. In comparing the prediction accuracy of the two learning approaches, we find that in this instance, the \textit{Ensemble} approach outperforms the \textit{Merged} by 15\%. Figure \ref{fig:Sup1} allows us to visualize where these improvements occur; while the \textit{Merged} is good at learning the true covariate-outcome at the densest regions of the covariate space (in this case, corresponding to the centers of the two clusters), the \textit{Ensemble} is able to additionally more closely learn the true model at the tails of the covariate distribution, akin to local smoothing. By focusing each forest on a single cluster, the \textit{Ensemble} method is able to focus more on all areas of the covariate space without getting pulled overly to the densest regions, while the \textit{Merged} concentrates on those with most representation within the training set. 

\end{appendix}

\end{document}